\documentclass[acmtog]{acmart}
\usepackage{booktabs}
\usepackage[ruled]{algorithm2e}
\usepackage{algpseudocode}
\usepackage{graphicx} 
\usepackage{subcaption}
\usepackage{tcolorbox}
\usepackage{listings} 
\usepackage{mdframed}
\usepackage{multirow}
\usepackage{makecell}

\usepackage{color, xcolor}
\usepackage{soul}
\soulregister\cite7
\soulregister\citep7
\soulregister\citet7
\soulregister\ref7
\soulregister\pageref7

\usepackage[skip=5pt]{caption}
\usepackage{afterpage}
\usepackage{mathtools}

\usepackage{zi4}  

\newcommand{\styledfileinput}[3][text]{%
  \begin{mdframed}[
    backgroundcolor=gray!10,
    linewidth=0pt,
    innertopmargin=2pt,
    innerbottommargin=2pt
  ]
    {\footnotesize\fontfamily{zi4}\selectfont
      \lstinputlisting[
        language=#1,
        basicstyle=\footnotesize\fontfamily{zi4}\selectfont,
        keywordstyle=\footnotesize\fontfamily{zi4}\selectfont,
        breaklines=true,
        tabsize=4,
        showstringspaces=false,
        showspaces=false,
        showtabs=false,
        frame=none,
        #3
      ]{#2}
    }
  \end{mdframed}
}

\SetAlFnt{\small}
\SetAlCapFnt{\small}
\SetAlCapNameFnt{\small}
\SetAlCapHSkip{0pt}

\acmJournal{TOG}
\acmVolume{44}
\acmNumber{6}
\acmArticle{}
\acmYear{2025}
\acmMonth{12}
\acmPrice{}

\setcopyright{rightsretained}

\acmDOI{10.1145/3763353}

\begin{document}
% Title portion
\title{Imaginarium: Vision-guided High-Quality 3D Scene Layout Generation}

\author{Xiaoming Zhu\textsuperscript{*}}
\orcid{0009-0002-3695-7288}
\affiliation{%
  \institution{Tsinghua University}
  \city{Shenzhen}
  \country{China}}
\email{zxiaomingthu@163.com}

\author{Xu Huang\textsuperscript{*}}
% \authornote{Equal contribution}
\orcid{0009-0006-2655-2251}
\affiliation{%
  \institution{Tencent}
  \city{Shenzhen}
  \country{China}}
\email{ydove1031@gmail.com}

\author{Qinghongbing Xie}
\orcid{0009-0000-9590-638X}
\affiliation{%
  \institution{Tsinghua University}
  \city{Shenzhen}
  \country{China}}
\email{xqhb23@mails.tsinghua.edu.cn}

\author{Zhi Deng\textsuperscript{†}}
% \authornote{Corresponding author}
\orcid{0000-0003-4582-4874}
\affiliation{%
  \institution{Tencent}
  \city{Shenzhen}
  \country{China}}
\email{zhideng@mail.ustc.edu.cn}

\author{Junsheng Yu}
\orcid{0009-0001-8719-1289}
\affiliation{%
  \institution{Southeast University}
  \city{Shenzhen}
  \country{China}}
\email{junshengyu33@163.com}

\author{Yirui Guan}
\orcid{0009-0006-0603-748X}
\affiliation{%
  \institution{Tencent}
  \city{Shenzhen}
  \country{China}}
\email{guan1r@outlook.com}

\author{Zhongyuan Liu}
\orcid{0000-0003-1601-0038}
\affiliation{%
  \institution{Tencent}
  \city{Shenzhen}
  \country{China}}
\email{lockliu@tencent.com}

\author{Lin Zhu}
\orcid{0009-0001-9044-9922}
\affiliation{%
  \institution{Tencent}
  \city{Shenzhen}
  \country{China}}
\email{hahnna0918@shu.edu.cn}

\author{Qijun Zhao}
\orcid{0009-0007-4797-9155}
\affiliation{%
  \institution{Tencent}
  \city{Shenzhen}
  \country{China}}
\email{qijunzhao@tencent.com}

\author{Ligang Liu}
\orcid{0000-0003-4352-1431}
\affiliation{%
  \institution{University of Science and Technology of China}
  \city{Hefei}
  \country{China}}
\email{lgliu@ustc.edu.cn}

\author{Long Zeng\textsuperscript{†}}
% \authornote{Corresponding author}
\orcid{0000-0002-3090-6319}
\affiliation{%
  \institution{Tsinghua University}
  \city{Shenzhen}
  \country{China}}
\email{zenglong@sz.tsinghua.edu.cn}

\thanks{$^*$Equal contribution, $^\dagger$Corresponding author}

\renewcommand\shortauthors{Zhu, Huang, et al.}

\begin{abstract}
Generating artistic and coherent 3D scene layouts is crucial in digital content creation. Traditional optimization-based methods are often constrained by cumbersome manual rules, while deep generative models face challenges in producing content with richness and diversity. Furthermore, approaches that utilize large language models frequently lack robustness and fail to accurately capture complex spatial relationships. To address these challenges, this paper presents a novel vision-guided 3D layout generation system. 
We first construct a high-quality asset library containing 2,037 scene assets and 147 3D scene layouts. Subsequently, we employ an image generation model to expand prompt representations into images, fine-tuning it to align with our asset library. We then develop a robust image parsing module to recover the 3D layout of scenes based on visual semantics and geometric information. Finally, we optimize the scene layout using scene graphs and overall visual semantics to ensure logical coherence and alignment with the images. Extensive user testing demonstrates that our algorithm significantly outperforms existing methods in terms of layout richness and quality. The code and dataset will be available at \url{https://github.com/HiHiAllen/Imaginarium}.
\end{abstract}




\begin{teaserfigure}
\centering
  \includegraphics[trim=0 15 0 15, clip, width=0.97\textwidth]{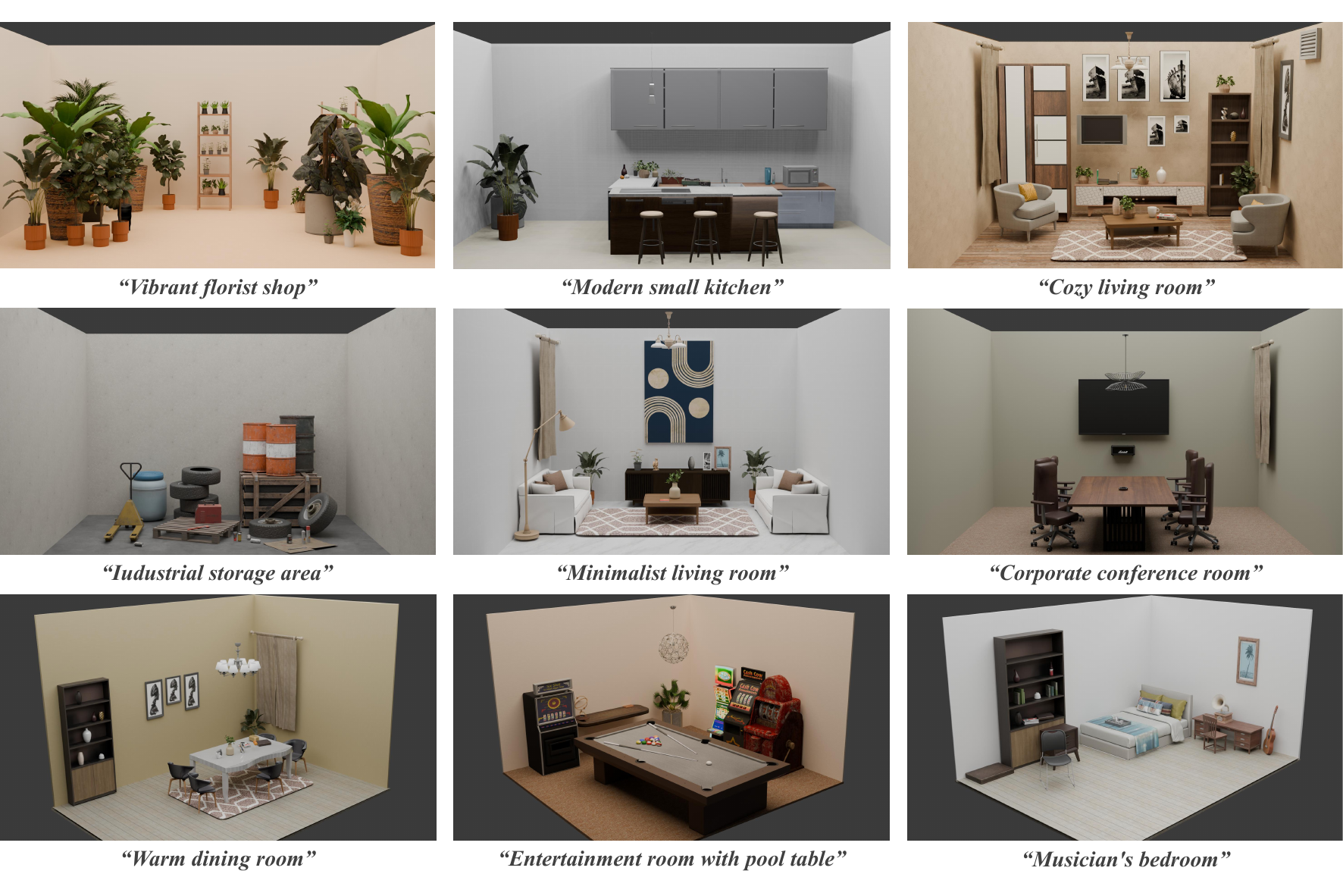}
  \caption{Some high-quality 3D scene layouts generated by our vision-guided system not only exhibit strong performance in indoor environments but can also be extended to outdoor scenes. The complete text prompts are provided in Appendix~\ref{subsubsec:Complete_scene_generation_prompts}.}
  \label{fig:teaser}
\end{teaserfigure}

\maketitle

\section{Introduction}~\label{introduction}
Generating logically coherent and visually appealing customized scene layouts from predefined asset collections presents significant challenges in digital content creation. This issue is particularly critical in fields such as game scene generation and computer-generated imagery (CGI) for films.

Traditional methods~\cite{yeh2012synthesizing,merrell2011interactive,chang2014learning,chang2017sceneseer,fisher2010context,Jiang_2018} frame this as a complex graph-based optimization problem, sampling from pre-modeled layout distributions and iteratively optimizing using predefined scene priors (e.g., layout guidelines, object category distributions). However, defining precise rules is both time-consuming and requires substantial artistic expertise. Furthermore, predefined rules may limit the expression of complex and diverse scene combinations.

More recent deep generative approaches~\cite{Wang_Yeshwanth_Niesner_2021,DBLP:conf/nips/PaschalidouKSKG21,nie2023learning,tang2024diffuscene} learn layout generators from pre-constructed 3D scene layout datasets. However, due to the high costs, privacy concerns, and time-consuming nature of collecting 3D data, these datasets remain relatively limited, leading to outputs that lack diversity and fail to meet the practical needs of artistic experts. This scarcity is particularly pronounced in new game or film productions, where preparing numerous diverse, high-quality 3D scene layouts in advance is nearly impossible, limiting the applicability of generators trained on native 3D data. While large language model-based scene generation methods~\cite{feng2024layoutgpt,yang2024holodeck,aguinakang2024openuniverseindoorscenegeneration} have emerged by extracting layout priors from language models and optimizing them with scene logic rules, they fundamentally lack spatial intuition and geometric precision, struggling to accurately represent complex spatial relationships, model object poses, and adhere to aesthetic design principles, ultimately limiting their effectiveness in creating realistic and coherent layouts.

Moreover, existing asset libraries like Objaverse~\cite{deitke2024objaverse} and 3D Future~\cite{fu20203dfuture3dfurnitureshape}, are often constrained by poor mesh quality, limited stylization options, and a heavy reliance on composite assets (e.g., a bookshelf with ornaments treated as a single asset), which restricts layout flexibility. To address these limitations, we curated a high-quality collection of 2,037 indoor and outdoor assets, which professional artists used to create 147 high-quality scene layouts—a dataset we plan to open-source to benefit the research community.

Recent advancements in image generation, driven by the explosive growth of image data and progress in diffusion-based models~\cite{ho2020denoisingdiffusionprobabilisticmodels,SahariaCSLWDGLA22ImageGen,ruiz2023dreambooth}, have significantly enhanced 2D generative capabilities. Building upon these developments and the substantial progress in foundational visual models \cite{liu2023matcher,yang2024depth,liu2025segment}(e.g., detection, segmentation, and depth estimation), we developed a visual-guided 3D scene layout generation system. This system is designed to transfer the rich and controllable generative capabilities of 2D image models to the task of 3D layout generation.

Our pipeline first utilizes the image generation model Flux~\cite{flux2024} to expand a user-input prompt into a guided image. After fine-tuning with our high-quality scene layout data, Flux generates images of higher quality that are also more consistent with the asset collection. Subsequently, we construct an image analysis module based on a pre-trained visual model, which integrates visual semantic segmentation, geometric parsing from a single image, and a graph-based scene graph logic construction module. Next, we adopt a semantic feature matching strategy to retrieve objects from the asset collection that are most similar to the guidance image. We then iteratively solve for the rotation, translation, and scaling transformations corresponding to each foreground object based on a combination of visual semantic features, geometric information, and scene layout logic. Finally, we perform consistency optimization on the overall 3D scene layout using scene graph logic and image semantic parsing, ensuring that the final scene layout is visually and logically close to the guided image.

Image generation models excel at producing aesthetically pleasing and detailed 2D layouts, and our approach leverages these capabilities for 3D scene layout tasks. Unlike previous methods that often rely on rigid composite assets (e.g., treating "a bowl of fruit on the table" as a single object), which leads to redundancy and insufficient diversity, our approach positions objects in varied poses and placements based on the guidance image. Furthermore, we introduce an internal layout function that allows assets to be arranged within other assets, optimizing space usage and improving scene realism. These capabilities result in more natural, detailed, and visually appealing 3D scene layouts. Experimental results show significant improvements in layout quality compared to previous methods.

In summary, our contributions are as follows:
\begin{itemize}
\item We have developed an innovative visual-guided system for high-quality scene layout generation.
\item We have established a high-quality 3D scene layout dataset, which will be open-sourced for community benefit.
\item We propose a robust scene object pose estimation algorithm integrating visual semantics with geometric information.
\end{itemize}
\section{Related Work} 

\subsection{Data-Driven Scene Layout Generation}
Data-driven scene layout generation methods fall into two main categories. The first employs manually defined scene priors and classical graphical models, optimized through non-linear optimization \cite{chang2014learning, fisher2012example, qi2018human, xu2013sketch2scene, yu2011make} or manual interaction \cite{chang2017sceneseer, merrell2011interactive, savva2017scenesuggest}. These priors follow design guidelines \cite{merrell2011interactive, yeh2012synthesizing}, object frequency distributions \cite{chang2014learning, chang2017sceneseer}, or human activity spaces \cite{fisher2015activity, fu2017adaptive, jiang2012learning, qi2018human, ma2016action}. While effective, this approach is limited by the time-intensive nature of manual prior design and model expressiveness constraints.

Recently, with advances in deep learning and improved 3D scene datasets \cite{fu20203dfuture3dfurnitureshape}, research has shifted toward end-to-end generators. Various approaches have emerged, including Spatial And-Or Graphs \cite{Jiang_2018}, autoregressive models \cite{paschalidou2021atiss, Wang_Yeshwanth_Niesner_2021, wang2018deep, nie2023learning}, 3D GANs \cite{yang2021indoor}, and Variational Autoencoders (VAEs) \cite{Yang_Zhang_Yan_Huang_Ma_Zheng_Bajaj_Huang_2021, Purkait_Zach_Reid_2020, Yang_Guo_Zhou_Tong_2021}. Despite offering quality improvements, these methods struggle with diversity, stability issues, and realism \cite{xiao2021tackling}. Recent diffusion-based models \cite{tang2024diffuscene, dahnert2024coherent} have enhanced layout richness by encoding object attributes (e.g., object categories, 6D poses, and textual descriptions from predefined asset libraries) in latent space. Building on this, InstructScene \cite{lin2024instructscene} first learns a scene-graph prior with a graph neural network (GNN) and uses it as the conditioning signal for the diffusion process, further improving layout fidelity and global coherence. Another line of work \cite{wald2020learning,dhamo2021graph,zhai2023commonscenes,zhai2024echoscene} model scene graphs from datasets and learn a generative distribution over them; at inference time, a scene graph is first generated and then used to reconstruct the corresponding 3D scene. However, these approaches remain limited by scarce 3D scene data, leading to overfitting and generalization challenges. Our method addresses these limitations by leveraging pretrained image generation models \cite{flux2024} to reconstruct 3D layouts from 2D images, significantly improving scene generation diversity.

\subsection{Language-Driven Scene Layout Generation}
The advent of large language models (LLMs) \cite{touvron2023llama, brown2020languagemodelsfewshotlearners, achiam2023gpt} has enabled textual-to-spatial scene synthesis through code interfaces. Pioneering works like HOLODECK \cite{yang2024holodeck} leverage LLMs to predict object categories, sizes, and positions via geometric constraints, while LayoutGPT \cite{feng2024layoutgpt} generates CSS-formatted layouts through chain-of-thought prompting. I-Design \cite{ccelen2024design} introduces multi-agent LLM collaboration. SceneCraft \cite{hu2024scenecraft} treats an LLM as an agent that authors Blender scripts, which are then executed to synthesize the 3D scene. However, these LLM-based methods often exhibit instability and artifacts, such as providing only four discrete pose estimation options, and face inherent limitations in scene complexity and aesthetics. Recent multimodal approaches show promising directions. Fireplace \cite{huang2025fireplace} renders 3D scenes as images to equip VLMs with 3D reasoning, thereby planning how objects are arranged. \cite{deng2025global} represents the scene as a hierarchical tree and uses a VLM to plan 3D object placements in a top-down manner. ARCHITECT \cite{wang2024architect} synergizes language guidance with diffusion models via hierarchical 2D inpainting to generate more detailed layouts, while LayoutVLM \cite{sun2024layoutvlm} combines vision-language models with differentiable optimization for physically valid layouts. Recent advances like CAST \cite{yao2025cast} reconstruct 3D scenes by generating individual objects and predicting poses through point cloud alignment with generative model representations. However, such approaches overlook the reusability of industrial assets with predefined properties beyond geometry, such as animations and interactive attributes.
While these methods demonstrate improved visual-semantic alignment, their reliance on fixed orientations and hard relational constraints for asset placement often leads to unnatural poses. Furthermore, the mismatch between arbitrarily generated image content and the available set of 3D assets creates a domain adaptation problem, resulting in final placements that significantly differ in style from the reference images. In contrast, our method directly extracts scene layout knowledge from visual models, leveraging style-consistent image guidance and continuous pose estimation to generate more natural and aesthetically pleasing scenes. The integration of scene graphs and geometric constraints further enhances system stability.

\subsection{Pose Estimation of Novel Objects}
Novel object pose estimation has evolved through geometric and learning approaches. Early works like PPF \cite{drost2010model} used geometric hashing, later enhanced by CNN features \cite{sundermeyer2020multi}. CAD alignment approaches emerged with Mask2CAD \cite{kuo2020mask2cad}, followed by ROCA \cite{gumeli2022roca}, SPARC \cite{langer2022sparc}, and DiffCAD \cite{gao2024diffcad}, which improved alignment through coordinate regression, iterative rendering, and diffusion modeling, respectively. However, their reliance on specific CAD libraries inherently limits open-set applicability. Complementary template matching methods achieve enhanced robustness with unseen objects by operating solely in the 2D domain. \cite{nguyen2022templates} applied CNN features for rotation estimation, while \cite{thalhammer2023self} demonstrated Vision Transformers' superiority in this task. MegaPose \cite{labbe2022megapose} employed a Coarse2Fine optimization strategy on a massive dataset, effectively generalizing to unseen objects. Building on this, FoundPose \cite{DBLP:conf/eccv/OrnekLTMKFH24} combined DINOv2 features with efficient template matching. Recently, GigaPose \cite{DBLP:conf/cvpr/NguyenGSL24} integrated template matching with local features, enhancing speed and robustness by fine-tuning DINOv2 through contrastive learning on the BOP challenge dataset.

\begin{figure*}[ht!]
    \centering
    \includegraphics[trim=30 110 30 100, clip, width=1\textwidth]{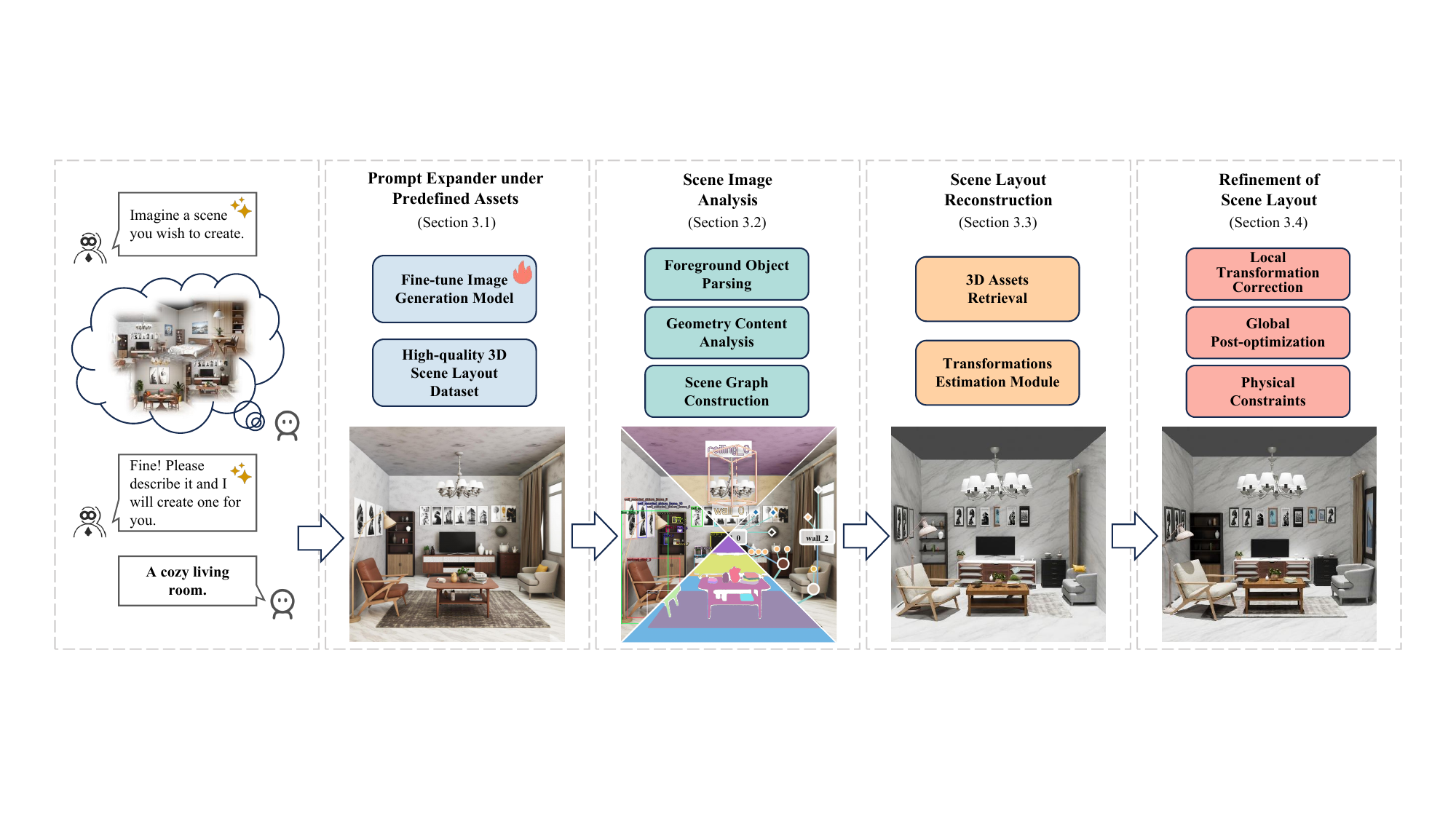}
    \caption{Overview of our method. We first transforms a text prompt into a detailed 2D guide image using a fine-tuned model, ensuring stylistic consistency with our asset library. This image is then analyzed for semantic, geometric, and relational information, guiding the retrieval, transformation estimation, and optimization of 3D assets into the final, coherent layout. See Appendix~\ref{subsubsec:supp_vis} for additional visualizations of intermediate steps.}
    
    \label{fig:pipeline}
\end{figure*}

In our task, the discrepancy between predefined assets and image content complicates pose estimation. We address this by utilizing GigaPose's finetuned DINOv2 \cite{nguyen2024gigapose} for category-based template rotation estimation, enhanced with geometric constraints and scene logic to ensure global consistency.

\section{Method}\label{sec:method}
\paragraph{Problem Statement} 
We aim to generate high-quality 3D scene layouts from a predefined set of 3D assets $A$ based on a user prompt. Mathematically, this is defined as a function $G$ that generates 3D layouts as follows:

\begin{equation}
\label{eq:problem_defined}
G(O|\text{prompt}, A) = \{o_1, o_2, o_3, \dots, o_n, \cdots\},
\end{equation}
where $\text{prompt}$ is a textual description (e.g., "the boss's office"). Each $o_i$ consists of $\{ \text{obj}_i, R_{i}, t_i, s_i \}$, where $\text{obj}_i$ is an asset from $A$ (geometry and texture), $R_{i} \in SO(3)$ is the rotation, $t_i \in \mathbb{R}^{3}$ is the translation, and $s_i \in \mathbb{R}^{3}$ is the scale of the asset.

\paragraph{Method Overview}  
The proposed vision-guided 3D scene layout generation system, shown in Fig.~\ref{fig:pipeline}, consists of three key stages. In Sec.~\ref{sec:method:prompt_to_image_content_expander}, we create a high-quality 3D scene dataset from $A$ and fine-tune the Flux-model to generate images that align with the stylistic characteristics of $ A $ and established design practices. In Sec.~\ref{sec:method:scene_image_parsing}, we develop a scene image analysis module that integrates visual semantic segmentation, geometric analysis, and scene graph construction. In Sec.~\ref{sec:method:3D-layout Reconstruction}, we use semantic feature matching to retrieve assets from $A$ that match the guiding image. We then estimate the rotation, translation, and scaling transformations of foreground objects based on visual and geometric data. Finally, in Sec.~\ref{sec:method:logical correction}, we refine these transformations through scene graph constraints and physical optimization to ensure a plausible 3D layout.

\subsection{Prompt Expander under Predefined Assets}\label{sec:method:prompt_to_image_content_expander}
\paragraph{Fine-tune Image Generation Model} 
Given a prompt input, we aim to generate 2D scene images that capture visual characteristics of a predefined 3D assets $A$, serving as guides for 3D scene layout reconstruction. Generating images that align with the style of $ A $ will robustly enhance visual asset retrieval and layout transformation estimation in later stages. To address limited view challenges, we focus on axonometric and frontal views for their comprehensive spatial coverage and design convention alignments. Following DreamBooth~\cite{ruiz2023dreambooth}, we use a unique tag $[\text{V}]$ to identify scene data, enabling efficient Flux model fine-tuning with minimal high-quality 3D layout renderings. We constructed a high-quality 3D scene dataset based on asset library $A$ for fine-tuning and evaluating. Our experiments reveal that the fine-tuned generated model as a prompt-to-scene expander trained on scenes built with $A$: it acquires consistent global patterns (viewpoint, rendering style) and moderate object-level features (textures, shapes), while maintaining creative layout flexibility. The visual similarity between objects in generated scenes and those in asset library $A$ effectively enhances visual asset retrieval and layout transformation estimation in subsequent stages.

\begin{figure*}[t!]
\centering
\includegraphics[trim=40 280 40 30, clip, width=1\textwidth]{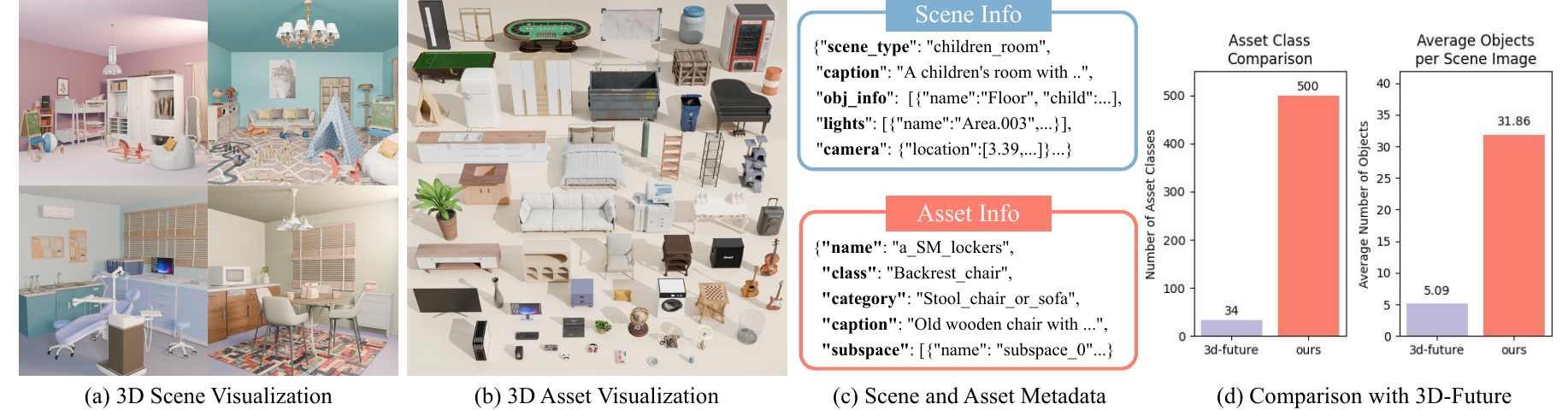}

\caption{Overview of our high-quality 3D scene layout dataset: (a) Representative 3D scenes with interior layouts. (b) Diverse 3D assets from our collection. (c) Structured metadata schema for scenes and assets. (d) Comparison with 3D-Future, highlighting our dataset's superior variety and complexity.}

\label{fig:dataset_intro}
\end{figure*}

\paragraph{High-quality 3D scene layout dataset}
We have developed a comprehensive 3D scene layout dataset that addresses critical limitations in existing resources, such as the prevalence of composite assets and limited variety in 3D-Future \cite{fu20203dfuture3dfurnitureshape}, and the stylization issues and low-quality models in Objaverse \cite{deitke2024objaverse}. As shown in Fig.~\ref{fig:dataset_intro}, it comprises 2,037 high-quality 3D models across 500 classes and 237 categories, with realistic textures and materials. These assets have been used to create 147 expertly designed scene layouts across 20 different types. Compared to 3D-Future, our dataset offers significantly higher asset diversity (500 classes vs. 34) and scene complexity (31.86 objects per scene vs. 5.09), enabling the creation of diverse, complex, and realistic scenes for both indoor and outdoor environments. These scenes were rendered into images for fine-tuning generative models.

The dataset was meticulously curated from a combination of custom-commissioned models, high-quality open-source content, and licensed marketplace items, which were then arranged into cohesive scenes by 20 professional artists with over three years of experience. To maximize its utility, the dataset is accompanied by comprehensive, multi-level annotations. At the asset level, annotations include descriptive captions and bounding box dimensions, with the crucial addition of manually annotated internal, placeable subspaces for assets that can contain other objects. At the scene level, we provide detailed scene captions, the spatial transformation matrix for each object (including the camera), parent-child hierarchical relationships, segmentation maps with individual object masks, and depth maps. Finally, all scenes were rendered using carefully positioned cameras to capture optimal axonometric and frontal viewpoints, ensuring maximum information content for subsequent 3D reconstruction tasks. A full statistical breakdown and visual examples are provided in Appendix~\ref{subsec:More_Details_for_Dataset}.

\subsection{Scene Image Analysis}\label{sec:method:scene_image_parsing} 
We utilize the prompt expander described in Sec.~\ref{sec:method:prompt_to_image_content_expander} to transform the prompt into a more expressive scene image $I$. Subsequently, we need to analyze the content of the image, which includes the semantic segmentation map of the foreground objects $S_{\text{fg}}$, the geometric proxy models for each object in the image, specifically the 3D oriented bounding boxes (OBBs) of the foreground objects, plane detection for walls, floors, and ceilings, as well as the logical relationships among the objects depicted in the scene.

\paragraph{Foreground Objects Semantic Parsing}
First, Based on the Chain of Thought (CoT) strategy \cite{wang2022towards}, we design a prompt incorporating predefined asset library categories (see Appendix~\ref{subsubsec:Prompt_for_Object_Extraction}) and input it with the image into GPT-4o to parse objects in the image. We transform these categories into a format suitable for grounding-dino detection through a category merging map $\mathcal{M}$, converting $\{\text{cate}_{i}^{A}\}$ into $\{\text{cate}_{i}^{g}\} = \{\mathcal{M}({\text{cate}_{i}^{A}})\}$. Using grounding-dino-1.5~\cite{ren2024grounding}, we obtain 2D bounding boxes $\{\text{bbox}_{i}^{2D}\}$, which we input into SAM~\cite{kirillov2023segment} to generate foreground segmentation results $S_{\text{fg}} = \{\mathbf{m}_{i}\}$.

\paragraph{Geometry Content Analysis}
We employ Depth Anything V2 \cite{yang2024depth} to estimate the depth map $D$ of the scene image and convert it to a point cloud $P$ using camera intrinsics $K$. For foreground regions $S_{\text{fg}} = \{\mathbf{m}_{i}\}$, we extract corresponding point clouds $\{P^{\mathbf{m}_{i}}\}$ and fit oriented bounding boxes (OBBs) $\{\text{obb}_{\mathbf{m}_{i}}\}$. For background regions, we apply RANSAC \cite{fischler1981random} to identify perpendicular planes representing walls, floor, and ceiling, by minimizing the Hausdorff distance between these planes and background points while enforcing orthogonality constraints.

\begin{figure}[ht!]
    \centering
    \includegraphics[width=0.48\textwidth]{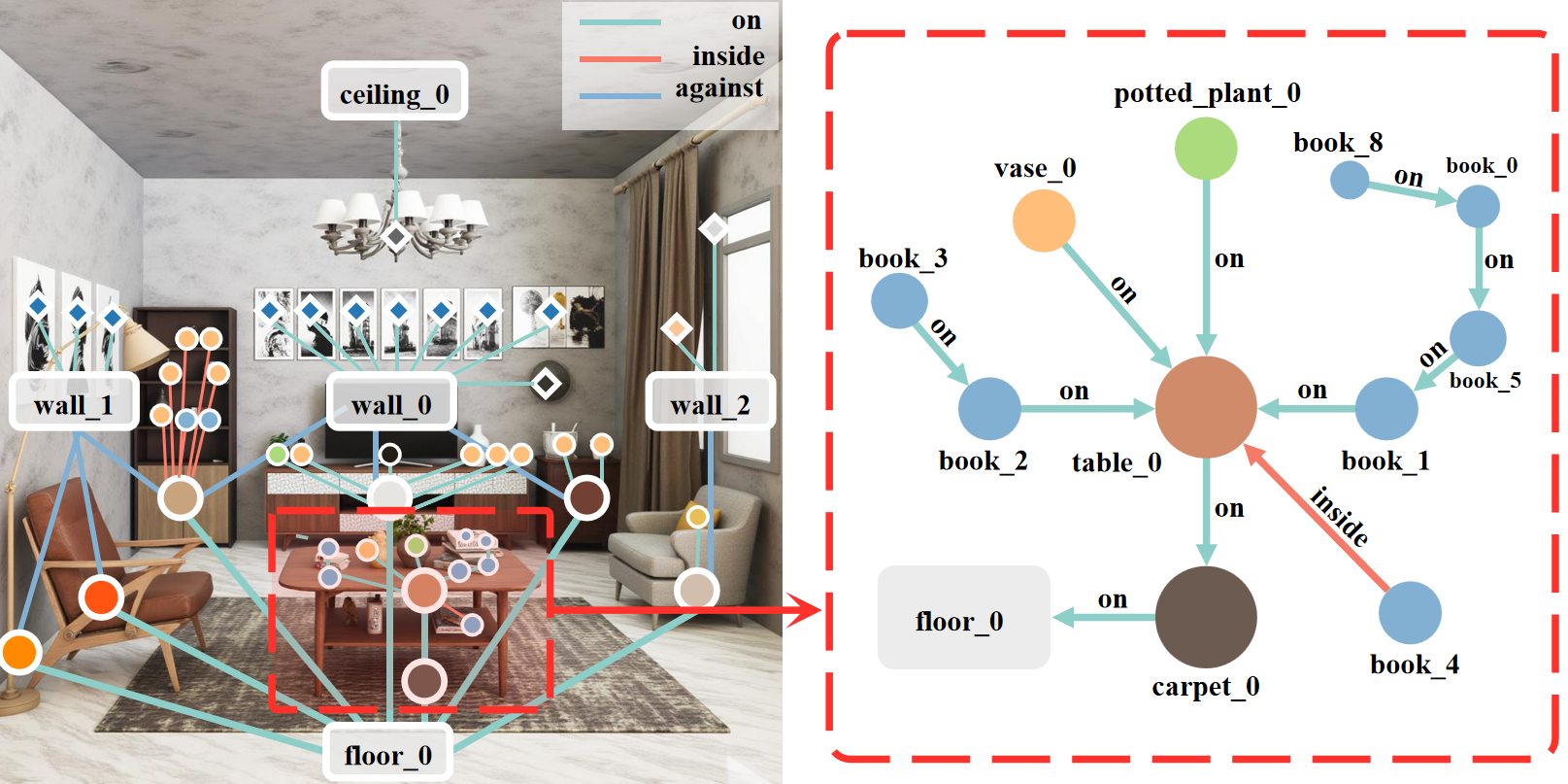}
    \makebox[0.235\textwidth]{\footnotesize(a)}\makebox[0.235\textwidth]{\footnotesize(b)}
    \caption{(a) Scene graph constraints extracted by our algorithm. (b) Close-up of the support relationship tree structure (highlighted in red box in (a)).}
    \label{fig:scene_graph}
    % \vspace{-0.3cm}
\end{figure}

\paragraph{Scene Graph Construction}
Based on multimodal model capabilities, we selected two key geometric relationships as shown in Fig.~\ref{fig:scene_graph}, which are easily interpreted from images and generalize well, even in quasi-outdoor scenes: (1) \textbf{Support Relationship}: Object $\text{obj}_{\text{a}}$ supports $\text{obj}_{\text{b}}$ ($\text{obj}_{\text{a}} \prec \text{obj}_{\text{b}}$) when $\text{obj}_{\text{b}}$ is positioned above $\text{obj}_{\text{a}}$, suspended by a ceiling, or contained within $\text{obj}_{\text{a}}$; and (2) \textbf{Wall Proximity Relationship}: Object $\text{obj}_{\text{b}}$ has contact with structural elements (walls, ceilings), defined as $d(\text{obb}_{\text{b}}, (n^{w}, t^{w}))=0$.

We construct the scene graph through a three-step process: (1) Analysis of the Floor Support Tree Structure using GPT-4o to determine floor-supported objects and establish a recursive support tree $\mathcal{T}$ with vertical relative distances $d^{\text{vertical}}$; (2) Analysis of Ceiling-Supported Objects; and (3) Analysis of Objects Against Walls, determining which objects contact specific walls. Detailed implementation of this procedure is provided in Appendix~\ref{subsubsec:scene_graph_construction}. Due to occlusions causing incomplete depth maps, we refine OBBs using above scene graph logical relationships. For objects supported by the floor, we ensure their OBBs maintain perpendicular relationships with the floor plane and extend them to make proper contact.

\subsection{Scene layout Reconstruction}\label{sec:method:3D-layout Reconstruction}
After analyzing the scene image, we reconstruct the scene layout corresponding to the predefined asset set $A$ through asset retrieval and transformation estimation based on visual features and geometric semantics to obtain the coarse scene layout.

\subsubsection{3D Asset Retrieval}
For each masked region $I_{\textbf{m}_{i}}$, we retrieve the most suitable 3D asset $\text{obj}_{\textbf{m}_{i}}$ from our assets library by combining inverse category mapping $\mathcal{M}^{-1}$ with visual feature similarity and size compatibility metrics (see Appendix~\ref{subsubsec:3d_asset_retrieval_details}).

\subsubsection{Transformations Estimation Module}
We first design a multi-step strategy based on visual features and geometric semantics to estimate the rotation transformations corresponding to the 3D assets $\{\text{obj}_{\textbf{m}_{i}}\}$. Then, we infer a coarse translation transformation based on the centers of $\{\text{obb}_{\textbf{m}_{i}}\}$. Finally, while ensuring that the deformation of the assets remains visually coherent, we maximize the intersection volume between $\text{obb}_{\textbf{m}_{i}}$ and $\text{obb}_{\text{obj}_{\textbf{m}_{i}}}$ to obtain the corresponding scale transformation for $\text{obj}_{\textbf{m}_{i}}$.

\paragraph{Rotation Transformation Estimation}
We employ a coarse-to-fine strategy combining visual semantics and geometric information:

\textit{Visual-semantic based candidates.} Following works~\cite{labbe2022megapose, nguyen2024gigapose}, we first render the asset \( \text{obj}_{\mathbf{m}_i} \) from 162 pre-sampled viewpoints $ V = \{v_{k}\}_{k=1}^{162} $ as $\mathcal{R}(\text{obj}_{\mathbf{m}_i}, v_k))$, and then extract pose-sensitive features \( F_{\text{ae}}(\mathcal{R}(\text{obj}_{\mathbf{m}_i}, v_k))_{\text{img}} \) using the feature extractor \( F_{\text{ae}}(\cdot)_{\text{img}} \) from GigaPose~\cite{nguyen2024gigapose}, which excels at detecting rotations perpendicular to the image plane (i.e., in-plane rotations). 
Finally, we establish the similarity to measure the overall similarity between the two images through the matching relationship of these features. The similarity is computed as follows:
\begin{equation}
\small
\label{eq:feature_similarity}
\text{sim}_{\text{img}}(I_{\textbf{m}_{i},v_{k}}^{A}, I_{\textbf{m}_{i}}) \!=\! \!{\sum_{j \in \mathcal{K}^{v_{k}}} \!\cos \left\langle \!F_{\text{ae}}(I_{\textbf{m}_{i},v_{k}}^{A})_{\text{img}}(j), F_{\text{ae}}(I_{\textbf{m}_{i}})_{\text{img}}(j) \!\right\rangle}
\end{equation}

where \( I_{\mathbf{m}_i, v_k}^A = \mathcal{R}(\text{obj}_{\mathbf{m}_i}, v_k) \) and \( \mathcal{K}^{v_k} \) represents the set of matching points determined by semantic feature similarity.

\textit{Coarse selection.} We aim to select the top \( k \) candidate views \( V_{\text{can}} \) from the 162 sampled viewpoints, focusing on views with higher keypoint correspondences and stronger semantic similarity. The top \( k \) candidate views are selected based on the feature similarity $\text{sim}_{\text{img}}(\cdot, \cdot)$. In our experiments, \( k \) is set to 10, ensuring the optimal view is among the candidates.

% \begin{figure}[t]
\begin{figure}[ht!]
\centering
\includegraphics[trim=5 0 5 0, clip, width=0.48\textwidth]{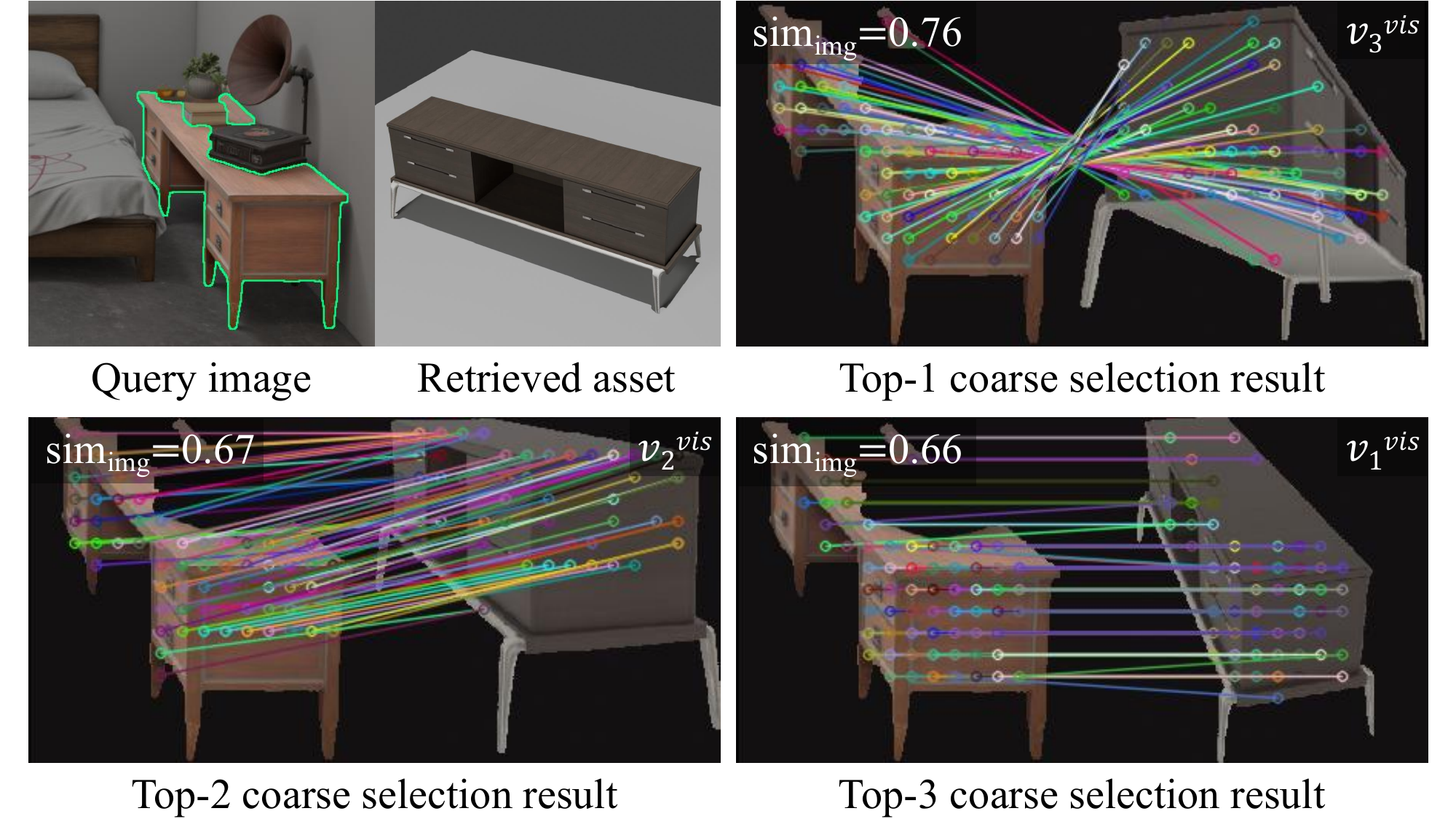}
\caption{Coarse-to-fine view selection. Coarse selection ranks views by keypoint match quality, while fine selection uses homography transformation to identify the most viewpoint-similar match (selecting $v_{1}^{\text{vis}}$ in this example).}
\label{fig:homograph_case}
\end{figure}

\textit{Fine selection.} For each candidate \(v_i\in V_{\text{can}}\), we first compute the homography transformation matrix \(H_v\) between the candidate view \(I^{\text{obj}}_{v_i}\) and the input image \(I_{\mathbf{m}_i}\) by RANSAC. We then analyze the difference between the homography transformation $H_{v}$ and the identity matrix, as in Eq.~\ref{eq:homograph}, this homography transformation analysis effectively suppresses errors in correspondences arising from symmetrical ambiguities (Fig.~\ref{fig:homograph_case}). The final top \(k=4\) views are those with the smallest Frobenius norm:

\setlength{\abovedisplayskip}{-1em}
\setlength{\belowdisplayskip}{3pt}
\setlength{\abovedisplayshortskip}{-1em}
\setlength{\belowdisplayshortskip}{3pt}

\begin{equation}  % 减小公式上方间距
    \label{eq:homograph}
    \{v_{i}^{\text{vis}}\}_{i=1}^{k} = \arg\min_{v\in V_{\text{can}}}^{(k)} \| U_{v}V_{v}^{T} - \text{I} \|_{F}^{2},
\end{equation}

where \( H_v = U_v \Sigma V_v^T \) is the singular value decomposition of \( H_v \), and \(\| \cdot \|_F\) denotes the Frobenius norm.

\begin{figure}[ht!]
\centering
\includegraphics[trim=200 75 200 70, clip, width=0.48\textwidth]{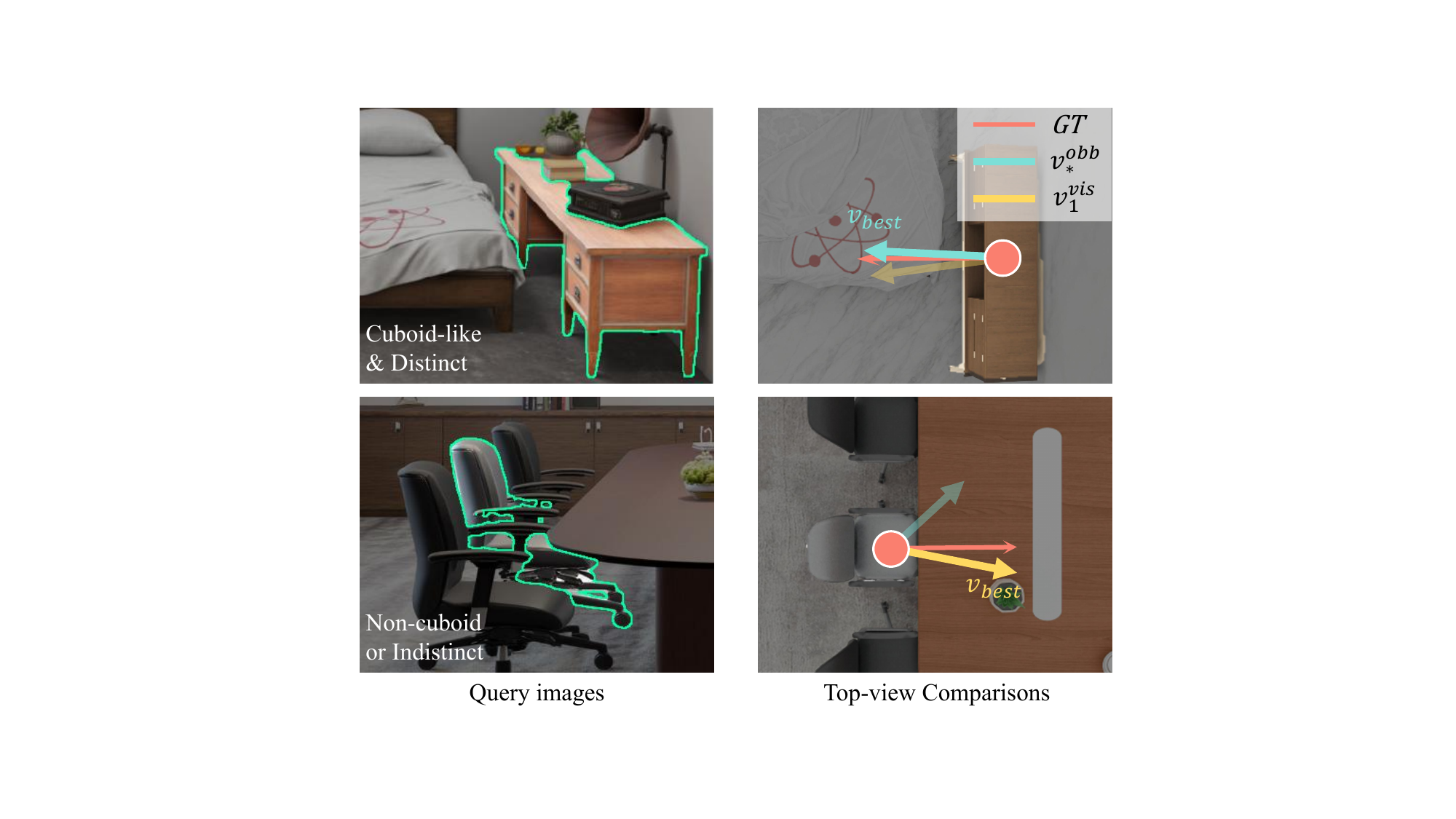}
\caption{Top-view illustration of candidates' geometric enhancement. Each row compares orientation estimations for different query scenarios, showing ground truth (GT), OBB-based ($v_{\ast}^{\text{obb}}$), and vision-based ($v_1^{\text{vis}}$) estimations. The best estimation ($v_{\text{best}}$) is highlighted, demonstrating the adaptive integration of geometric guidance.}
\label{fig:top_view_rotation_comparsion}
\end{figure}

\textit{Geometric enhancement of candidates.} Leveraging the geometric consistency from single-image depth recovery, particularly with cuboid-like assets, we refine rotation estimation using the accurate OBBs obtained in Sec.~\ref{sec:method:scene_image_parsing}. For well-defined cuboid objects, the four orientations of the OBB's vertical planes, \( \{v_{i}^{\text{obb}}\}_{i=1}^{4} \), guide the rotation transformation \( \{[R^{v_{i}^{\text{obb}}}]\}_{i=1}^{4} \). However, for non-cuboid shapes or incomplete point clouds due to occlusions or errors, we use an adaptive strategy to ensure robustness. The final rotation is selected by minimizing the angular difference between candidate viewpoints and geometric viewpoints:

\begin{equation}
\label{eq:final_rotation_selection_1}
    (v_{\ast}^{\text{obb}}, v_{\ast}^{\text{vis}}) = \arg\min_{\substack{v^{\text{obb}} \in \{v_{i}^{\text{obb}}\},\\ v^{\text{vis}} \in \{v_{j}^{\text{vis}}\}}} \arccos\left(\frac{\text{Trace}( R^{v^{\text{vis}}}{}^T  R^{v^{\text{obb}}}) - 1}{2}\right)
\end{equation}
% }
$$
    v_{\text{best}} = \left\{
    \begin{array}{ll} 
        v_{\ast}^{\text{obb}}, & \quad \text{if} \quad \theta \leq \tau, \\
        v_1^{\text{vis}}, & \quad \text{if} \quad \theta > \tau.
    \end{array}
    \right.
$$

Here, \( v_{\text{best}} \) is the selected viewpoint, $\theta$ is the angle between the view $v_{\ast}^{\text{obb}}$ and $v_{\ast}^{\text{vis}}$, and \( \tau = \frac{\pi}{5} \) in our experiments. This approach prioritizes OBB-based estimation for cuboid-like objects and defaults to \( v_1^{\text{vis}} \) when the OBB guidance is unreliable (Fig.~\ref{fig:top_view_rotation_comparsion}).

\begin{figure}[t]
    \centering
    \includegraphics[trim=20 0 20 0, clip, width=0.38\textwidth]{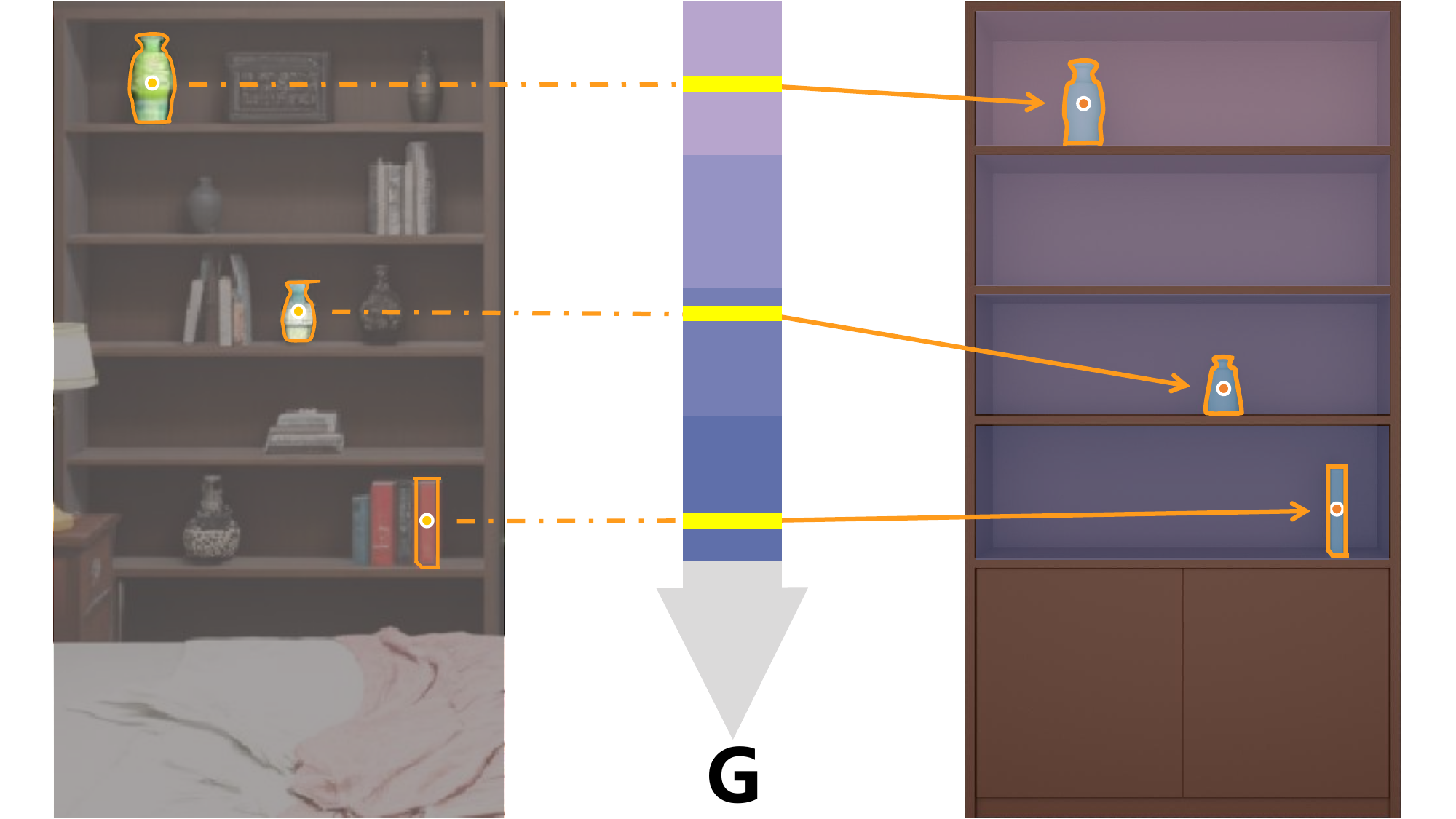}
    \caption{Internal placement logic illustration. Left: query object (yellow outline). Right: internal subspaces of the target container. Objects are placed in the nearest subspace based on the vertical distance $d^{\text{vertical}}$ between the centers of the object and the subspace along the gravity direction(G).}
    \label{fig:inner_placement}
\end{figure}

\paragraph{Translation and Scale Transformation Estimation}
For translation, we begin by approximating object positions using the OBB centers. For scaling, we optimize asset dimensions according to the object type: vertically adjustable, slender objects with two principal axes, or fully scalable objects. This ensures both practical placement and the preservation of each asset’s inherent design integrity (more details in Appendix~\ref{subsubsec:scale_transformation_details}).

\subsection{Refinement of Scene Layout}\label{sec:method:logical correction}
After individually estimating transformations for foreground objects, ambiguities may arise from depth errors and asset discrepancies. We resolve these through a novel three-stage refinement: optimizing rotation and scale using scene graph relationships, formulating constrained optimization for translations that preserves visual alignment while enforcing physical constraints, and applying physics simulation to ensure realistic object behaviors.

\subsubsection{Local Transformation Refinement based on Scene Graph}
We first optimize rotation and scale transformations using scene graph constraints. For rotation, we align object OBBs with their supporting surfaces, ceiling, or walls based on logical relationships in the scene graph. The support tree $\mathcal{T}$ enables recursive adjustment of rotational transformations following parent-child relationships. For objects placed inside containers (Fig.~\ref{fig:inner_placement}), we perform scale adjustments based on container capacity. When $\text{obj}_{\textbf{m}_{j}}$ is internally supported by $\text{obj}_{\textbf{m}_{i}}$ with vertical distance $d^{\text{vertical}}_{\textbf{m}_{j} \prec \textbf{m}_{i}} > 0$, we identify the pre-compute internal subspace and resize $\text{obj}_{\textbf{m}_{j}}$ accordingly.

\subsubsection{Global Post-optimization for Translation Transformations}
We optimize object positions to ensure physical plausibility through a constrained formulation that enforces non-intersection between objects, proper support hierarchies, ceiling attachments, and wall proximity requirements. we construct an objective function balances adherence to initial positions with visual segmentation alignment, while satisfying spatial constraints derived from the scene structure:

\vskip -1em
\begin{equation}
\small
\label{eq:refined_pose_via_logstic}
\begin{aligned}
    & \min_{\{t^{\text{update}}_{i}\}}  \quad \sum_{i} \lambda_{1} \| t_{i} - t_{i}^{\text{update}} \|_2^2 + \|\textbf{m}_{i} - \mathcal{R}_{\textbf{m}}(\text{obj}_{\textbf{m}_i}, v_{\text{ref}})\|_2^{2}. \\
    \text{s.t.} & \left\{
    \begin{array}{l} 
        \text{obj}_{\textbf{m}_i} \cap \text{obj}_{\textbf{m}_j}  = \emptyset, \quad \text{ if } i \neq j, \\
 z(\text{obj}_{\textbf{m}_i})_{\text{max}} = t^{c}, \quad \text{ if } i \in \text{C, Supported by Ceiling},\\
         d(\text{obj}_{\textbf{m}_i}, \text{obj}_{w})=0, \quad \text{ if }\text{obj}_{\textbf{m}_i} \text{is against } \text{obj}_{w},\\
          z(\text{obj}_{\textbf{m}_j})_{\text{min}} = z(\text{obj}_{\textbf{m}_i})^{*} ,\quad \text{ \!\!\!\!\!if }\text{obj}_{\textbf{m}_i} \text{and }  \text{obj}_{\textbf{m}_j} \text{meet } \mathcal{T}.
    \end{array}
    \right.
\end{aligned}
\end{equation}

Here, we set $\lambda_{1} = 0.1$ in our experiments. The variables $t_{i}$ and $t_{i}^{\text{update}}$ represent the initial and optimized positions of the object $\text{obj}_{\textbf{m}_i}$, respectively. The function $\mathcal{R}_{\textbf{m}}(\cdot, \cdot)$ renders the geometry of the object to obtain a mask image, where $v_{\text{ref}}$ denotes a reference viewpoint for depth conversion into a consistent point cloud, shared across all objects in the experiments. The values $z(\text{obj})_{\text{min}}$ and $z(\text{obj})_{\text{max}}$ denote the minimum and maximum $z$ values, respectively. $d(A, B) = \inf \{ \| a - b \| \mid a \in A, b \in B \}$. We solve this optimization in two steps: preprocessing support and wall constraints, then applying simulated annealing~\cite{SkiscimG83} using efficient voxel-based intersection calculations. Full details are in Appendix~\ref{subsubsec:Two_step_translation_optimization}.

\subsubsection{Physical Constraints}
Finally, we apply physical simulation using Blender's physics engine to ensure objects follow real-world physical behaviors, particularly important for elements like pillows on beds or stacked objects. More details are in Appendix~\ref{subsubsec:More_Details_in_Physical_Simulation}.

\begin{figure*}[ht!]
\centering
\includegraphics[trim=0 0 0 0, clip, width=1\textwidth]{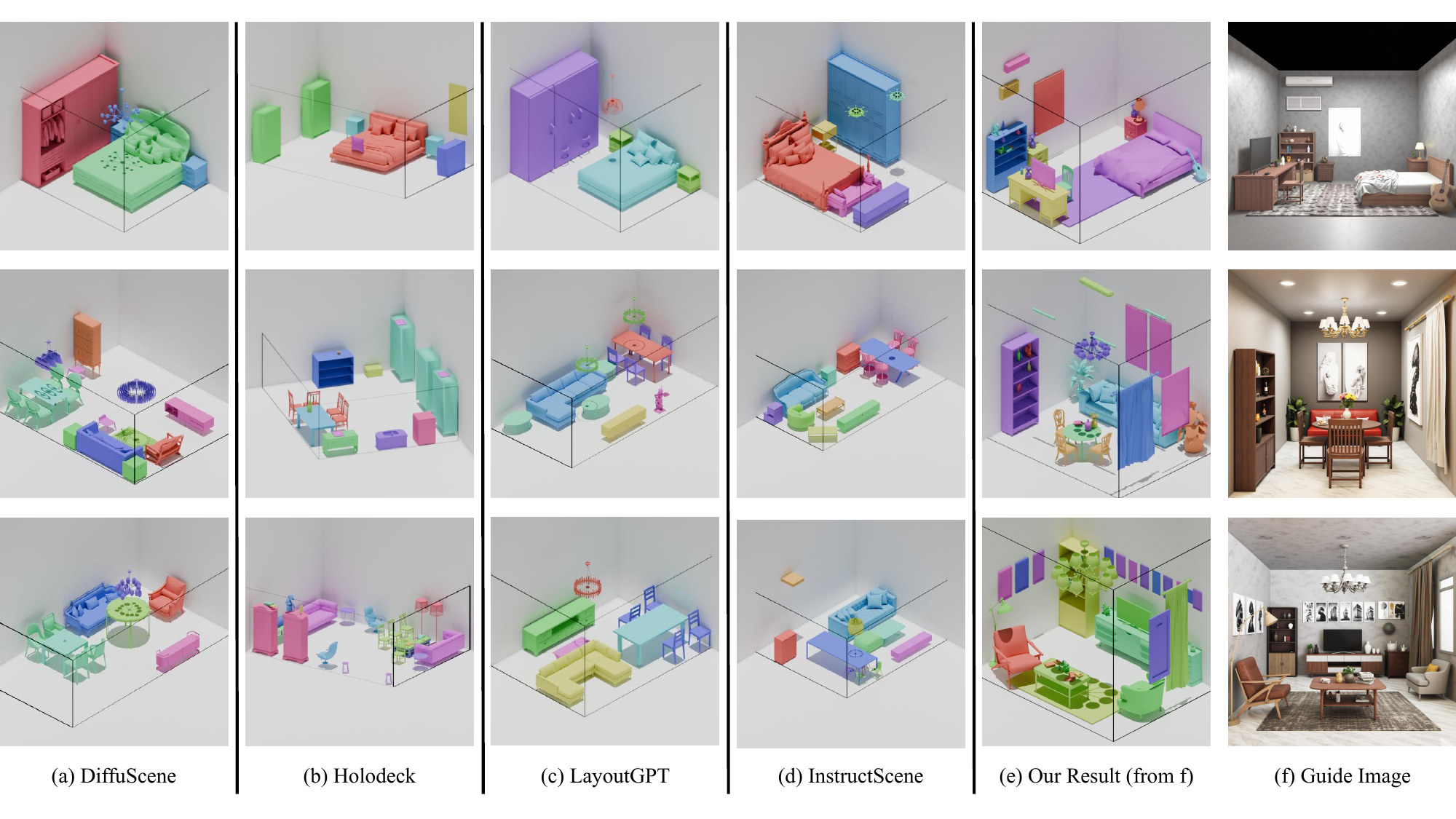}
\caption{Comparison of our generated 3D scene layouts with other state-of-the-art methods, illustrating the richness of our 3D generated layouts. More examples of our generated layouts are shown in Appendix~\ref{subsec:More_Qualitative_Results}.}
\label{fig:result_comparision}
\end{figure*}

\section{Experiments}\label{sec:exp}
We evaluate our system through comprehensive user studies and experiments focusing on: quality assessment, rotation estimation from single images, and ablation studies.

\subsection{Implementation Details}\label{sec:method:implementation_details}
We finetune the Flux model on our proposed dataset, which contains 147 unique scenes. The training data consists of images rendered with Blender at a resolution of 1024$\times$1024 pixels. To ensure a comprehensive representation of the scene layout, camera perspectives were manually selected, focusing on axonometric and frontal views. The training is conducted on two A100 GPUs for 15 epochs using LoRA with a rank of 16 and a learning rate of 1e-4. Following the DreamBooth~\cite{ruiz2023dreambooth} strategy, we employ a regularization technique that uses a unique identifier, \texttt{[V]}, for our in-domain data while including samples without this token for generalization.

Our system takes approximately 240 seconds to run on a single A100, with the following time distribution: text-to-image generation (10 seconds), scene image analysis (110 seconds), scene layout reconstruction (60 seconds), and layout refinement (60 seconds).

\subsection{Quality Assessment}
\subsubsection{Evaluation by Senior Art Students}
We invited 100 senior art students (ages 20-24) to evaluate our 3D scenes against HOLODECK \cite{yang2024holodeck}, LayoutGPT \cite{feng2024layoutgpt}, DiffuScene \cite{tang2024diffuscene}, and InstructScene \cite{lin2024instructscene}. These methods represent two layout generation strategies: LLM-guided approaches and data-driven generative models. For each method, we prepared 15 scenes per scene type (living room, dining room, and bedroom), totaling 45 scenes. Note that DiffuScene only supports these three scene types, while LayoutGPT is further limited to living room and bedroom scenes. For fair comparison, we removed all textures to focus on layout quality and standardized the asset database for Holodeck and LayoutGPT (we couldn't replace DiffuScene and InstructScene's assets due to its training-based nature requiring substantial data). Participants answered two questions:

\textbf{Q1:} Which layout appears more reasonable and realistic?

\textbf{Q2:} Which layout is more coherent and aesthetically pleasing?

\begin{table}[ht]
\centering
\caption{Comparison of preferable rates (\%) for different methods.}
\label{tab:user_study_senior_students}
\resizebox{0.5\textwidth}{!}{
\begin{tabular}{l|ccc|ccc}
\toprule
\multirow{2}{*}{\textbf{Our vs.}} & \multicolumn{3}{c|}{\textbf{Reasonable \& Realistic}} & \multicolumn{3}{c}{\textbf{Aesthetic}} \\
\cmidrule(lr){2-4} \cmidrule(lr){5-7}
 & \textbf{Dining} & \textbf{Living} & \textbf{Bedroom} & \textbf{Dining} & \textbf{Living} & \textbf{Bedroom} \\ 
\midrule
DiffuScene & 75.69 & 82.59 & 79.37 & 74.86 & 85.57 & 80.72 \\
Holodeck   & 79.27 & 77.08 & 76.79 & 82.72 & 72.92 & 74.55 \\
LayoutGPT  & -- & 76.69 & 76.50 & -- & 77.54 & 81.11 \\ 
InstructScene & 66.33 & 68.46 & 61.29 & 69.39 & 75.17 & 72.90 \\
\bottomrule
\end{tabular}
}
\end{table}

As shown in Table~\ref{tab:user_study_senior_students}, our method consistently outperforms all baselines. For reasonableness and realism, our approach achieves average preference rates of 79.22\%, 77.71\%, 76.60\%, and 65.36\% compared to DiffuScene, HOLODECK, LayoutGPT, and InstructScene respectively. For aesthetic quality, our method demonstrates even stronger advantages with preference rates of 80.38\%, 76.73\%, 79.33\%, and 72.49\% against the same competitors. Fig.~\ref{fig:result_comparision} provides visual comparisons of these results.

\subsubsection{Evaluation by Professional Artists on Richness}  \label{subsubsec:professional_eval}
We recruited 20 professional artists, each with at least three years of experience, to evaluate 60 scenes across three room types (Living Room, Dining Room, and Bedroom). The artists rated three dimensions—overall composition, semantic logic, and aesthetic appeal—on a 1-5 scale. To ensure a fair comparison with baseline methods, we conducted additional evaluations where textures were removed. These scenes were also evaluated by GPT-4o on the same dimensions. A score of 3 was set as the baseline, representing the average level compared to professionals. Detailed in Appendix~\ref{subsubsec:Prompt_for_GPT4_evalutation}.  

~\begin{table}[h]
\centering
\caption{Expert and GPT-4o evaluation comparison.}
\label{tab:expert_gpt4o_comparison}
\resizebox{0.49\textwidth}{!}{
\begin{tabular}{lcccc}
\hline
\textbf{Method} & \textbf{Composition} & \textbf{Semantic} & \textbf{Aesthetic} & \textbf{Overall} \\ \hline
\textbf{Ours} & 3.35/3.16 & 3.29/2.86 & 3.37/3.16 & 3.34/3.06 \\
DiffuScene & 2.86/3.07 & 2.80/2.78 & 2.83/3.07 & 2.83/2.97 \\
HOLODECK & 2.71/2.91 & 2.56/2.55 & 2.80/2.86 & 2.69/2.77 \\
LayoutGPT & 2.42/2.97 & 2.26/2.83 & 2.35/2.97 & 2.34/2.92 \\
InstructScene & 2.91/3.07 & 2.75/2.83 & 2.89/3.08 & 2.85/2.99 \\ \hline
\end{tabular}
}
\end{table}

As seen in Table~\ref{tab:expert_gpt4o_comparison}, our method consistently outperforms all baseline approaches, scoring 3.34 from human artists and 3.06 from GPT-4o, indicating performance on par with or slightly better than professional standards.

\subsubsection{Fidelity and Similarity of 3D Scene Layout Reconstruction}
We randomly selected 30 scenes from our dataset and used their rendered images to evaluate our system's reconstruction ability against ground-truth layouts. Objects supported by the ground or ceiling, or located near walls, were classified as primary objects crucial for scene structure, while others were considered secondary objects. Our evaluation includes seven key metrics: object recovery rates, category preservation rates, rotation AUC@60°, translation AUC@0.5m, scene graph relationship accuracy, CLIP similarity, and GPT-4o's assessment of layout fidelity.

Results in Table~\ref{tab:scene-evaluation-intra-set} show high fidelity in primary object recovery (92.31\%) and category preservation (95.83\%). The system also achieves strong geometric accuracy in rotation (74.83\% AUC@60°) and translation (84.32\% AUC@0.5m), along with 93.26\% scene graph accuracy. Secondary objects achieve lower recovery rates (70.41\%) due to resolution limitations and detection model constraints on smaller objects. CLIP similarity and GPT-4o ratings further confirm layout fidelity. Additional 3D scene layouts with their corresponding guide images are presented in Appendix~\ref{subsubsec:ablation_study_flux}.

\begin{table}[htbp]
\centering
\caption{Fidelity and layout similarity evaluation using dataset scenes.}
\label{tab:scene-evaluation-intra-set}
\setlength{\tabcolsep}{6pt}
\resizebox{0.48\textwidth}{!}{
\begin{tabular}{llcc}
\toprule
\multicolumn{2}{l}{\textbf{Metric}} & \textbf{Primary} & \textbf{Secondary} \\
\midrule
\multirow{5}{*}{\textbf{Fidelity}} & Object Recovery & 92.31\% & 70.41\% \\
& Category Preservation & 95.83\% & 91.67\% \\
& Rotation (AUC@60°) & 74.83\% & 71.51\% \\
& Translation (AUC@0.5m) & 84.32\% & 80.40\% \\
& Scene Graph Accuracy & \multicolumn{2}{c}{93.26\%} \\
\midrule
\multirow{3}{*}{\textbf{Similarity}} & CLIP (Guide Image) & \multicolumn{2}{c}{27.03} \\
& CLIP (Render Image) & \multicolumn{2}{c}{25.83} \\
& GPT-4o Rating & \multicolumn{2}{c}{8.29/10} \\
\bottomrule
\end{tabular}
}
\end{table}

\subsection{Comparison of Rotation Transformation Estimation}
We evaluate our rotation transformation estimation on the 3D-Future category asset pose estimation dataset, 3DF-CLAPE, which we derived from the 3D-Future dataset to better align with layout generation scenarios. It contains two subsets: (1) \textbf{3DF-CLAPE-Category} with 5,833 query-template pairs across 34 categories for category-level evaluation, and (2) \textbf{3DF-CLAPE-Instance} with 3,252 pairs for instance-level evaluation. Following standard practice \cite{wang2019normalized, Shotton_Glocker_Zach_Izadi_Criminisi_Fitzgibbon_2013, Li_Wang_Ji_Xiang_Fox_2020}, we report mean average precision (mAP) at various rotation error thresholds and the area under the curve (AUC).

Due to our unique task of open-set pose estimation for category-level CAD models from single images, we select several benchmarks that have shown potential in this domain: DINOv2, SPARC, and DiffCAD, AENet, GigaPose, Orient Anything \cite{wang2024orient}. 

\begin{table}[htbp]
\centering
\caption{Quantitative comparison of rotation estimation methods using AUC@60°. (OrientA: Orient Anything; GigaP: GigaPose)}
\label{tab:pose_result}
% \footnotesize
\resizebox{0.49\textwidth}{!}{
\setlength{\tabcolsep}{2pt}
\renewcommand{\arraystretch}{1.2}
\begin{tabular}{c|ccccccc}
\hline
\textbf{AUC@60°} $\uparrow$ & DINOv2 & SPARC & DiffCAD & OrientA & GigaP & AENet & \textbf{Ours} \\
\hline
Category-level & 31.68\% & 52.54\% & 26.45\% & 56.07\% & 39.85\% & 45.32\% & \textbf{70.06\%} \\
Instance-level & 31.38\% & 61.46\% & 25.44\% & 56.24\% & 57.43\% & 62.16\% & \textbf{81.44\%} \\
\hline
\end{tabular}
}
\end{table}

\begin{figure}[ht!]
    \centering
    \includegraphics[width=0.48\textwidth]{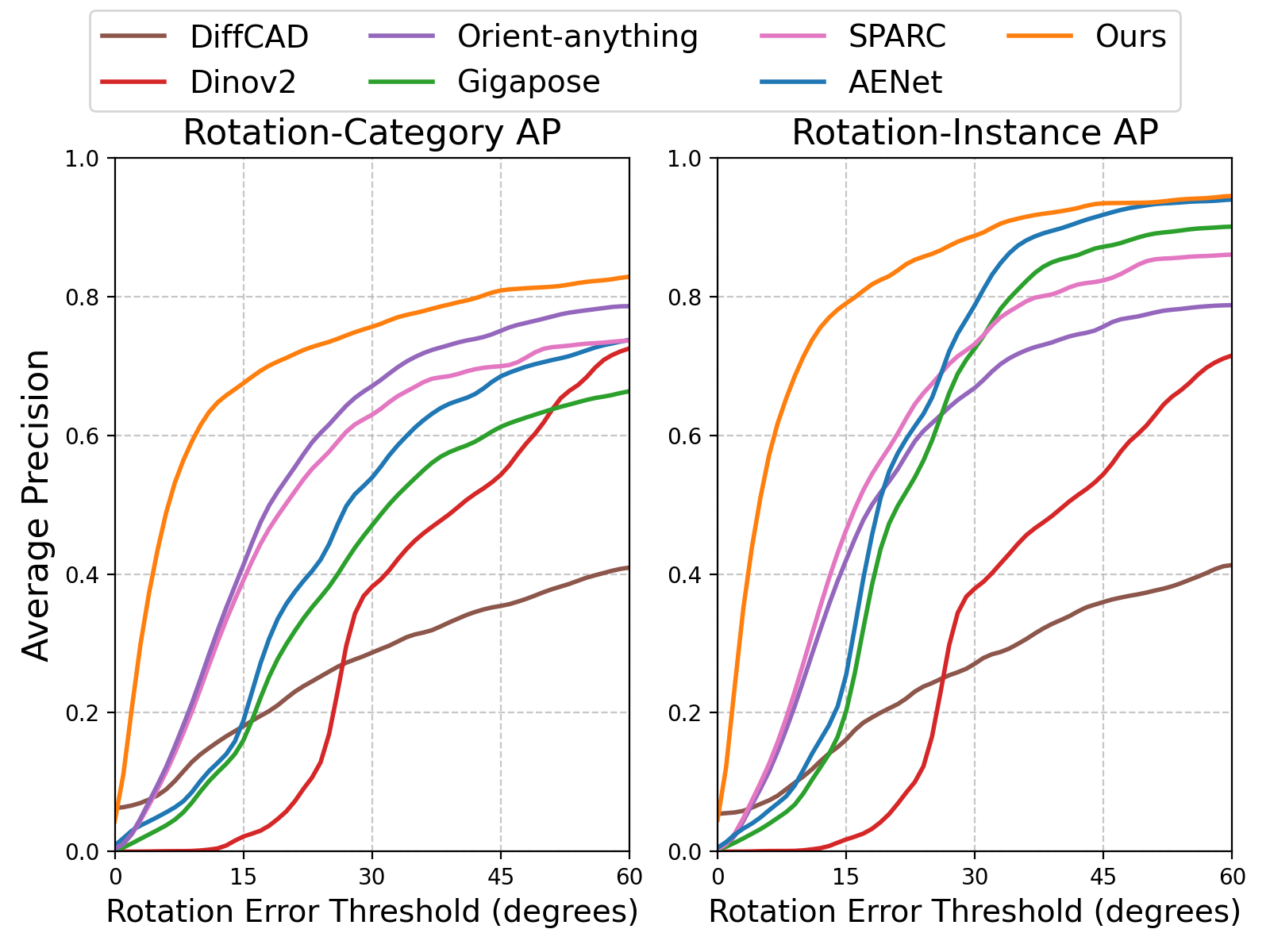}
    \caption{Comparison of performance in category and instance level rotation estimation with other methods.}
    \label{fig:overall_category_ap_curve_combined}
\end{figure}

As shown in Table \ref{tab:pose_result}, our approach achieves an AUC@$60^{\circ}$ of 70.06\% for category-level and 81.44\% for instance-level evaluation, significantly surpassing all benchmarks. Fig. \ref{fig:overall_category_ap_curve_combined} further demonstrates that our method outperforms existing approaches in category-level rotation estimation, achieving mAP values of 50.5\%, 65.5\%, and 80.5\% at thresholds of $5^{\circ}$, $15^{\circ}$, and $45^{\circ}$ respectively. Despite GigaPose using the same keypoint extractor (AENet) as our method, it underperforms due to limitations in handling template-query discrepancies. The results demonstrate both CAD-based approaches' superiority for 3D scene layout and the critical role of query-template similarity in pose estimation, shown by template matching methods outperforming non-template approaches (Orient Anything: 56.24\% AUC) and the marked improvements in instance-level tasks where query-template similarity is highest.

\subsection{Ablation Study}
We conduct comprehensive ablation studies to validate our key design choices across three components: (1) finetuning the Flux diffusion model, (2) rotation transformation estimation with homography and geometric information, and (3) scene layout refinement pipeline. These studies demonstrate that each component meaningfully contributes to system performance while maintaining generative diversity and physical plausibility.

\begin{figure*}[ht!]
    \centering
    \includegraphics[trim=10 10 10 0, clip, width=1\textwidth]{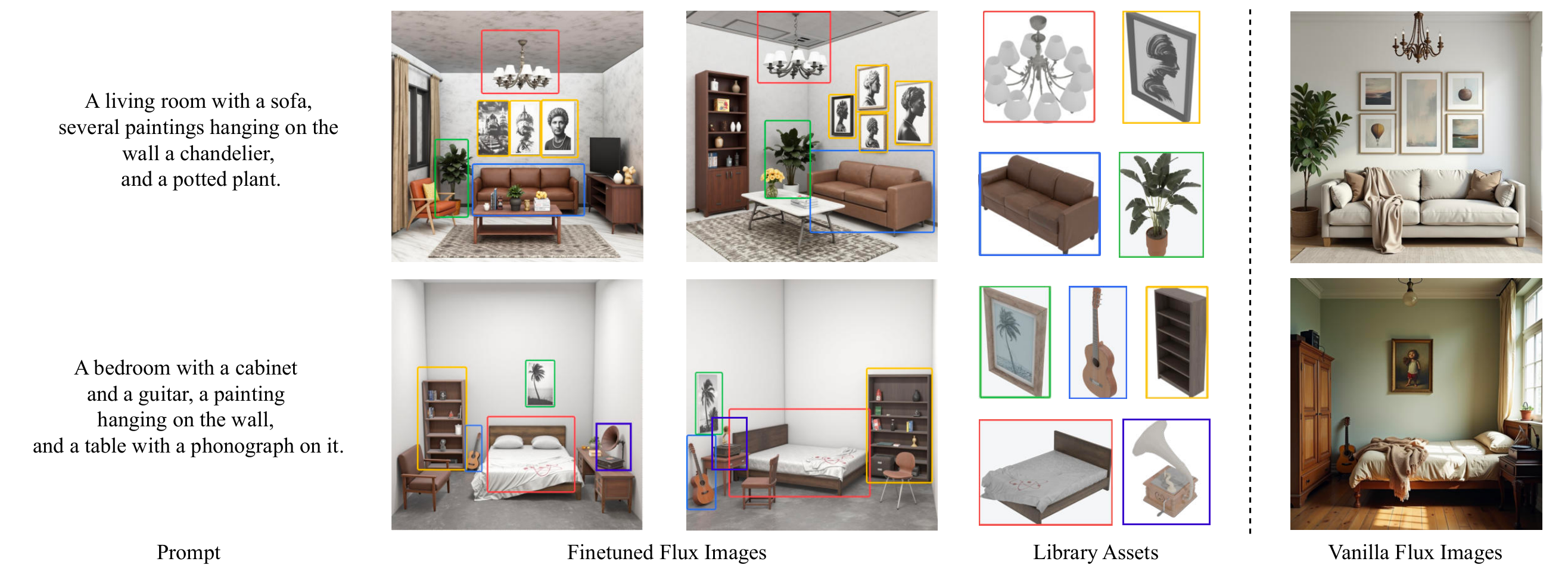}
    \caption{Comparison between Finetuned Flux and Vanilla Flux generated images. Given identical prompts (left column), Finetuned Flux (second column) generates images with objects that more closely resemble assets in our 3D library (third column), compared to Vanilla Flux (right column). This alignment improves retrieval accuracy and pose estimation, enabling more precise scene parsing and strengthening system robustness.}
    \label{fig:finetuned_flux_and_vanilla_flux}
\end{figure*}

\subsubsection{Ablation study of finetuned Flux} \label{subsubsec:ablation_study_flux}
We evaluate the impact of Flux finetuning through comprehensive ablation studies. While our system functions with off-the-shelf Flux, targeted finetuning enhances retrieval accuracy and pose estimation without sacrificing generative diversity. As shown in Fig.~\ref{fig:finetuned_flux_and_vanilla_flux}, the finetuned model generates images better aligned with our asset library given identical prompts. Table~\ref{tab:pose_result} demonstrates substantial pose estimation improvements (AUC@60° from 70.06\% to 81.44\%) when query objects match CAD models. We compare vanilla and finetuned Flux regarding retrieval accuracy, overfitting potential, and diversity preservation.

\paragraph{Retrieval Accuracy}
Based on the generation of 100 scene images each by Vanilla Flux and Finetuned Flux, we utilized our image analysis pipeline to identify 2343 objects and 2204 objects in the corresponding scenes, respectively. In addition, we manually identified the ground truth matches for these objects from our 3D asset library. The retrieval performance was evaluated using Top-1 and Top-3 accuracy:
 
\begin{table}[htbp]
\centering
\caption{Accuracy comparison between vanilla and finetuned models.}
\label{tab:accuracy_comparison}
\resizebox{0.45\textwidth}{!}{
\begin{tabular}{l@{\hspace{10mm}}c@{\hspace{10mm}}c}
\toprule
\textbf{Metric} & \textbf{Vanilla Flux} & \textbf{Finetuned Flux} \\
\midrule
\text{Top-1 Accuracy} & 48.57\% & 68.70\% \\
\text{Top-3 Accuracy} & 68.57\% & 83.21\% \\
\bottomrule
\end{tabular}
}
\end{table}

\begin{figure*}[ht!]
    \centering
    \includegraphics[trim=0 35 0 0, clip, width=0.95\textwidth]{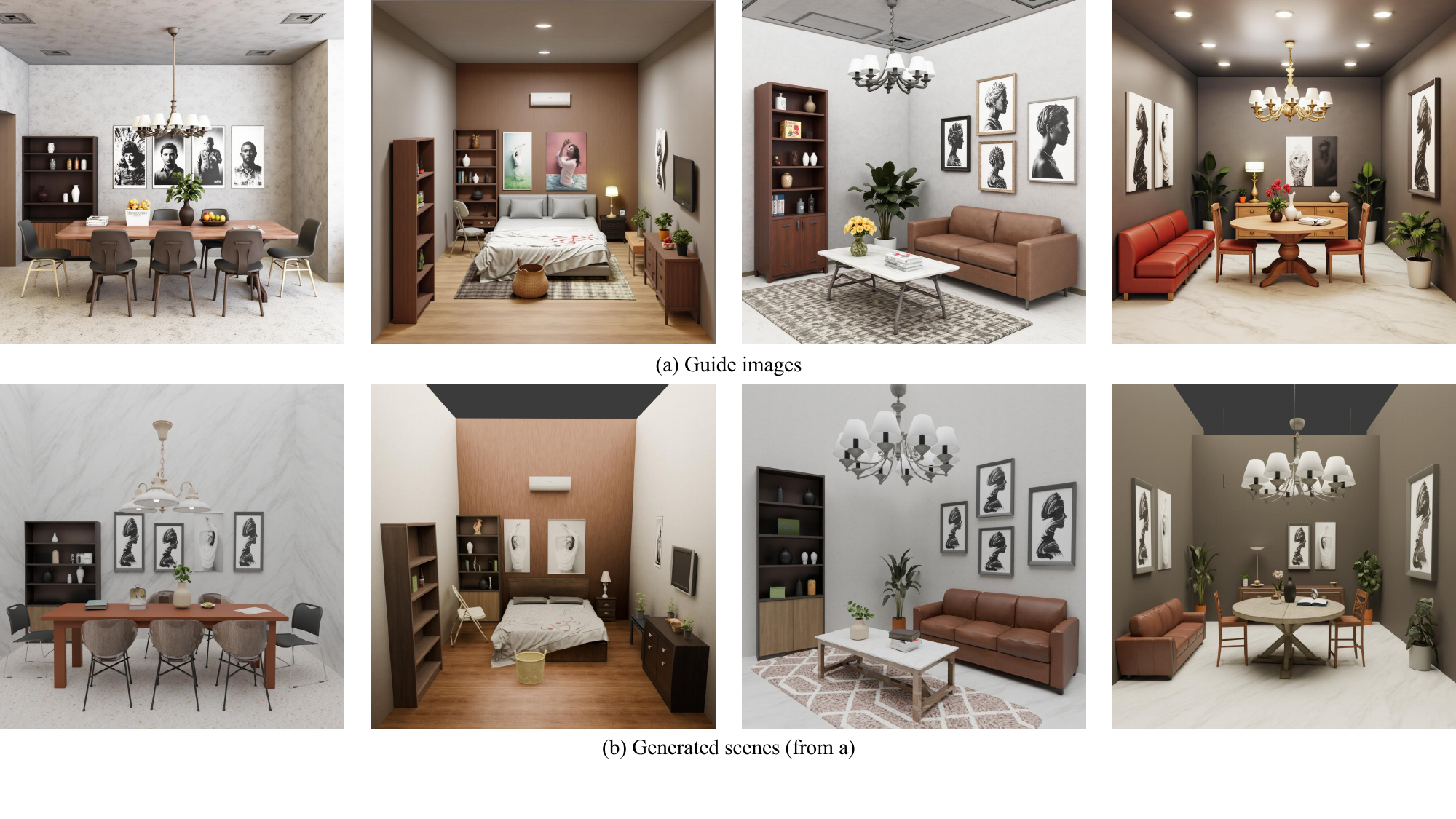}
    \caption{Additional results showcasing our method's ability to generate coherent 3D layouts from diverse guide images. The generated scenes (bottom) demonstrate high fidelity to the input's spatial arrangement and style.}
    \label{fig:gen_res_and_guide_img}
\end{figure*}

The substantial improvement demonstrates that finetuning enhances the model's ability to generate scenes aligned with our 3D asset library. More layouts with guide images, as show in Fig.~\ref{fig:gen_res_and_guide_img}.

\begin{table}[htbp]
\centering
\caption{Comparison of overfitting and diversity metrics.}

\label{tab:overfitting_metrics}
\resizebox{0.4\textwidth}{!}{
\begin{tabular}{lcc}
\toprule
\multirow{2}{*}[-0.5ex]{Model} & \multicolumn{2}{c}{Overfitting} \\
\cmidrule(lr){2-3}
 & NN LPIPS $\uparrow$ & Scene Sim. to Training $\downarrow$ \\
\midrule
Vanilla Flux & 0.6375 & 0.3665 \\
Finetuned Flux & 0.5981 & 0.3899 \\
\midrule
\multirow{2}{*}[-0.5ex]{Model} & \multicolumn{2}{c}{Diversity} \\
\cmidrule(lr){2-3}
 & DIV (LPIPS) $\uparrow$ & Intra-set Scene Sim. $\downarrow$ \\
\midrule
Vanilla Flux & 0.5782 & 0.2974 \\
Finetuned Flux & 0.5901 & 0.3178 \\
\bottomrule
\end{tabular}
}
\end{table}

\paragraph{Overfitting Evaluation}
To evaluate whether the Flux model is overfitting, we initially employed the Nearest Neighbor (NN) LPIPS distance to measure the visual similarity between the generated scene images and their closest matches in the training set. Additionally, following previous studies~\cite{henderson2017generative, ritchie2019fast}, we adopted a scene-to-scene similarity function to specifically assess the similarities in scene layouts (the detailed methodology is provided in Appendix~\ref{subsubsection:Scene_Layout_Similarity_Metric}). As shown in Table~\ref{tab:overfitting_metrics}, higher NN LPIPS values indicate less visual overfitting, while lower scene similarity scores suggest a reduction in layout overfitting. The finetuned Flux exhibits comparable NN LPIPS to the vanilla model, with only slightly higher scene similarity, indicating minimal overfitting. This confirms that our model generates novel arrangements rather than memorizing the training set. 
% Fig.~\ref{fig:layout_grid_vis} presents several example scenes.

\paragraph{Diversity Preserving}
Following DreamBooth we generated 20 images for each of 6 diverse prompts and calculated both visual diversity (DIV) using average pairwise LPIPS distances and layout diversity (Intra-set Scene Sim.) using average pairwise scene-to-scene similarity within each prompt set. Table~\ref{tab:overfitting_metrics} shows that the finetuned model maintains comparable visual and layout diversity to the vanilla model.

\paragraph{Learning Dynamics Analysis}
Our experiments reveal a clear learning hierarchy: the finetuned Flux model readily learns style and viewpoint (as visually apparent in Figs.~\ref{fig:result_comparision}, \ref{fig:finetuned_flux_and_vanilla_flux}, \ref{fig:gen_res_and_guide_img}, and \ref{fig:layout_grid_vis}), moderately captures object textures, but preserves layout diversity. We hypothesize this stems from varying supervision strengths—style and viewpoint provide strong global patterns across all training data, shapes and textures offer moderate signals through repeated object appearances, while layouts remain weakly learned due to scene uniqueness and multi-body constraints complexity. This hierarchical learning aligns with our goal of enhancing retrieval and pose estimation while maintaining generative flexibility.

\subsubsection{Ablation study of rotation transformation estimation}
\begin{table}[htbp]
\centering
\caption{Ablation study of our rotation transformation estimation module.}
\label{tab:ablation_study_rotation}
\resizebox{0.49\textwidth}{!}{
\begin{tabular}{ccc|cccc}
\toprule
AENet & Homography & Geometry & mAP@5 & mAP@15$^\circ$ & mAP@45$^\circ$ \\
\midrule
\checkmark &  &  & 4.30\% & 15.34\%  & 67.92\% \\
\checkmark & \checkmark &  & 5.21\% & 59.42\% & 76.07\% \\
\checkmark &  & \checkmark & 36.22\%  & 71.73\% & 77.16\% \\
\checkmark & \checkmark & \checkmark  & 66.57\% & 75.28\% & 80.61\% \\
\bottomrule
\end{tabular}
}
\end{table}
Table~\ref{tab:ablation_study_rotation} presents the ablation study of our rotation transformation estimation. The results highlight the significance of each component in our coarse-to-fine approach. The incorporation of homography significantly enhances performance(as show in Fig.~\ref{fig:homograph_case}), achieving mAP@$45^{\circ}$ of 76.07\% and mAP@$15^{\circ}$ of 59.42\%. Furthermore, the integration of geometric information further improves estimation accuracy, particularly at lower error thresholds, with mAP@$5^{\circ}$ increasing from 5.23\% to 36.22\%. Our complete model, which combines all components (AENet, homography, and geometry), achieves the best performance across all metrics. This demonstrates the effectiveness of our approach in combining visual-semantic features with geometric information for precise rotation estimation.

\subsubsection{Ablation study of scene layout refinement}
We evaluate the impact of each step in the scene layout refinement process, focusing on local refinement, global optimization, and physical constraints using three metrics: the support correctness rate, representing the percentage of correctly supported objects; the intersection pairs count, which quantifies geometric object collisions; and a GPT-4o evaluation that scores the overall aesthetic and logical quality of the scene following Sec.~\ref{subsubsec:professional_eval}. As shown in Table~\ref{tab:ablation_study_refinement}, each step contributes clear improvements, with global optimization playing the most critical role in fixing support relationships and reducing object interference while maintaining the scene's logical plausibility.

\begin{table}[htbp]
\centering
\caption{Ablation study of scene layout refinement.}
\label{tab:ablation_study_refinement}
\resizebox{0.48\textwidth}{!}{
\begin{tabular}{lccc}
\toprule
\textbf{Method} & \textbf{Supp. Corr. (\%)} & \textbf{Inter. Pairs} & \textbf{GPT-4o (1-5)} \\
\midrule
Initial Estimation & 62.45 & 5.43 & 2.83 \\
+ Local Refinement & 72.86 & 4.43 & 3.07 \\
+ Global Optimization & 90.80 & 2.21 & 3.26 \\
+ Physical Constraints & 91.34 & 2.20 & 3.29 \\
\bottomrule
\end{tabular}
}
\end{table}

\begin{figure}[ht!]
% \begin{figure}[H]
    \centering
    \includegraphics[trim=100 40 100 40, clip, width=0.48\textwidth]{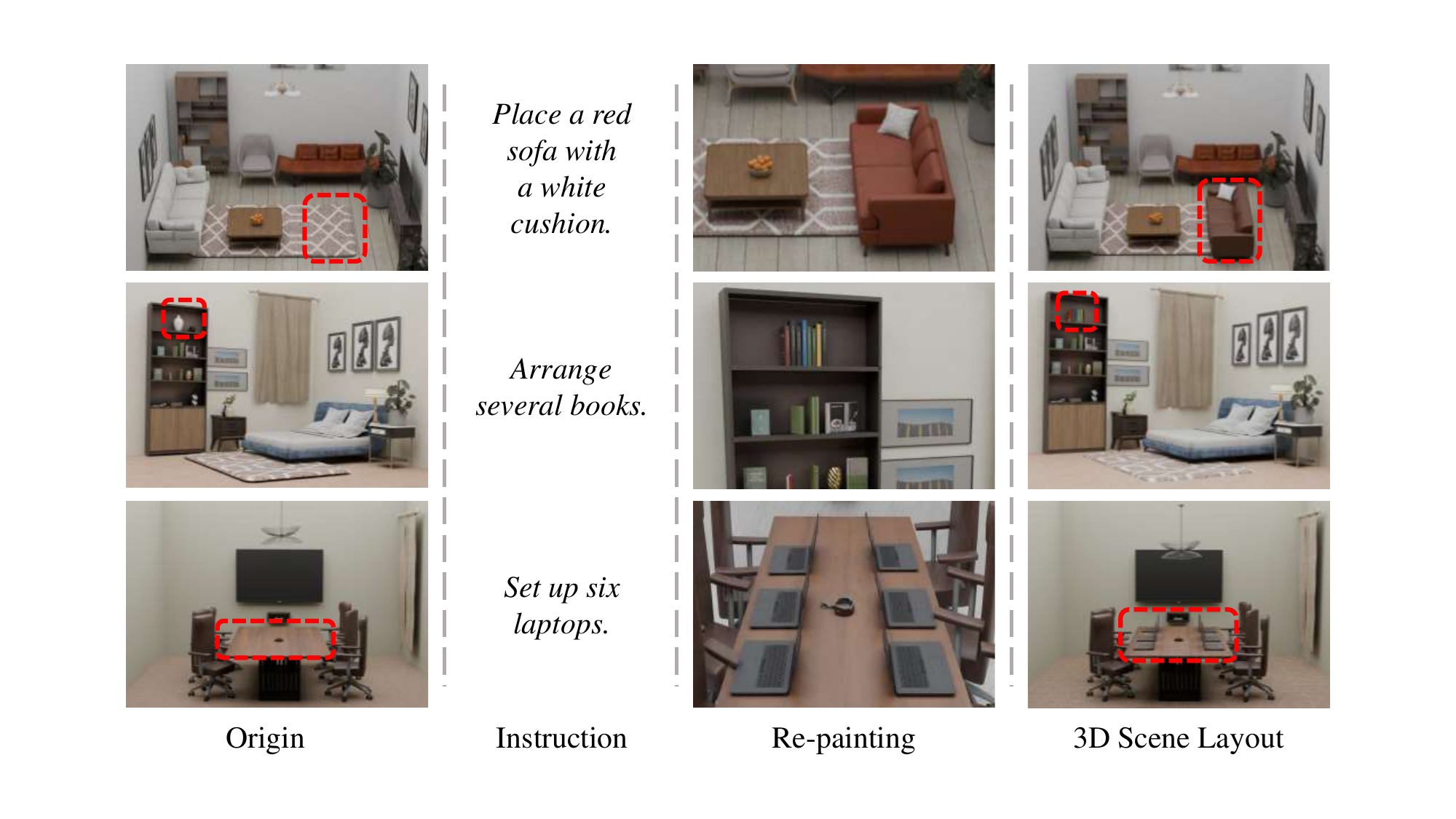}
    \caption{It showcases some re-editing examples that we generated using the Image Generation model. Using the text prompts from the second column, we re-paint the local information within the red box of the images in the first column using Flux, thereby controlling the 3D layout. This control over local information can achieve a very robust effect.}
    \label{fig:editing}
\end{figure}

\section{Application}

Generating 3D scenes typically requires significant time and expertise from skilled artists, making a straightforward method for re-editing essential. Unlike previous approaches based on large language models (LLMs) or 3D generation models, our method allows for more granular editing based on image manipulation techniques. As shown in Fig.~\ref{fig:editing}, we present several detailed editing examples, including global scene completion, object replacement, and local object addition. After generating the 3D scene, we can obtain renderings of both the global scene and any specific local area. By leveraging the capabilities of image generation models to fill in masked regions, we can perform fine-grained, controllable re-painting of any part of the scene, including specific objects and their exact positions. After re-painting, we fix the objects outside the masked area and utilize our algorithm to re-retrieve and estimate the relevant poses of the objects within the masked region.

\section{Conclusion and Discussion}
We present a visual-guided 3D scene layout system that generates coherent, aesthetic scenes from text or Canny images within 240 seconds, significantly reducing the 2.5-hour time typical in professional workflows. Our approach integrates Flux for layout generation, fine-tuned on our asset library for style consistency and more aligned asset selection. Unlike previous methods, we dynamically use image guidance for object orientations, creating more natural 3D layouts. User studies with 100 art students and 20 professional artists demonstrate significant performance advantages over current SoTA methods.

\textit{Limitation and Future Work.} 
While our approach achieves high-fidelity layouts, it is constrained by certain limitations. Despite our current progress, fine-tuning the image Generation model to achieve high consistency across multiple objects in complex scenes remains a primary challenge. Additionally, accurate pose estimation from single images remains challenging, particularly with severe occlusions. These failure modes are visually detailed in Appendix~\ref{subsec:failure_analysis}. We anticipate these limitations will diminish as visual foundation models advance. To specifically address pose ambiguity, incorporating multi-view perspective information from methods like MVD \cite{liu2024one, liu2024oneplus, liu2024syncdreamer} offers a promising path for more robust scene analysis. Looking forward, our system shows promise as an automated 3D data generation engine by transforming abundant 2D vision model placement knowledge into 3D asset placement data, addressing data scarcity in 3D scene generation tasks~\cite{ost2021neural, raistrick2024infinigen}. This enables more efficient training of models for 3D scene understanding and layout generation. Finally, introducing more coherent editing capabilities between 2D and 3D~\cite{xu2025sketch123, yan20243dsceneeditor,Deng_Sketch2PQ} is a meaningful exploration for making our future system more user-friendly.

\begin{acks}

This work was supported by the National Key R\&D Program of China (Grant No. 2022YFB3303101, 2022YFB3303400) and the National Natural Science Foundation of China (62025207).

We thank Yihang Wang for his efforts on the first demo and constructive suggestions on flux fine-tuning, and Zhi Ji, Yuxuan Xie, and Town Zhang for their helpful discussions and technical support.

\end{acks}

\bibliographystyle{ACM-Reference-Format}
\bibliography{main}

\clearpage
\appendix
\section{Supplementary Materials}

\subsection{Technical Implementation Details} \label{subsec:implementation_details}
\subsubsection{Scene Graph Construction} \label{subsubsec:scene_graph_construction}
Our scene graph construction involves extracting geometric logical relationships from foreground regions $S_\text{fg} = \{\textbf{m}_{i}\}$. Due to complex object shapes and occlusions, we combine qualitative image analysis using multimodal models with precise geometric methods. The process consists of three steps:

\textit{Step 1: Analysis of the Floor Support Tree Structure}.  
We build a tree-structured scene graph supported by the floor using a recursive approach, as illustrated in Algorithm~\ref{Alg:tree_floor_construction} and Algorithm~\ref{Alg:supported_relationship_analysis}. Using GPT-4o, we analyze each foreground object $\mathcal{F} = \{I_{\textbf{m}_{i}}\}$ through prompting to determine floor support. For floor-supported objects, we identify assets located within or above them through recursive geometric search, establishing a complete support tree $\mathcal{T}$ while retaining vertical relative distances $d^{\text{vertical}}$ for subsequent asset placement. Our experimental results show 91.95\% accuracy for this analysis.

\begin{algorithm}[ht!]
    \caption{Establishing Tree Node Relationships Supported by the Floor}
    \label{Alg:tree_floor_construction}
    \KwIn{Foreground object parsing from the scene image $\mathcal{F}=\{I_{\textbf{m}_{i}}\}$ and corresponding oriented bounding boxes $\text{obb}_{\textbf{m}_{i}}$;}
    \KwResult{A tree $\mathcal{T}$ representing relationships based on floor support.}

    Queue = \{\} \;

    \tcp{Identify floor-supported subnodes}
    \For{$\forall I_\textbf{m} \in \mathcal{F}$}{
        \tcp{Determined by GPT-4o prompt}
        \If{$\textbf{m}$ is supported by the floor}{
            AddLeafNode($\mathcal{T}$, floor, $I_\textbf{m}$)\;
            $\mathcal{T}(I_\textbf{m})[d^\text{vertical}] \gets 0$\\
            Queue.insert($I_\textbf{m}$)\;
        }
    }
    
    \tcp{Recursively constructing the support relationship tree}
    \While{!Queue.empty()}{
         $I_\textbf{m} \gets \text{Queue.pop()}$ \;
        \For{$I_{\textbf{m}_{n}} \in \mathcal{F}$}{
            \If{$d(\textbf{m}_{n}, \textbf{m}) < \epsilon$}{
                % $S_{\textbf{m}_{n} \prec \textbf{m}}$, $d_{\textbf{m}_{n} \prec \textbf{m}}^{\text{vertical}}$ as analyzed by supported relationship algorithm in Supp. Sec.~\ref{subsec:supported_relationship_analysis}.\\
                $S_{\textbf{m}_{n} \prec \textbf{m}}$, $d_{\textbf{m}_{n} \prec \textbf{m}}^{\text{vertical}}$ as analyzed by supported relationship algorithm~\ref{Alg:supported_relationship_analysis}.\\
                
                \If{$S_{\textbf{m}_{n} \prec \textbf{m}}$}{
                    AddLeafNode($\mathcal{T}$, $I_\textbf{m}$, $I_{\textbf{m}_{n}}$)\;
                    $\mathcal{T}(I_\textbf{m})[d^\text{vertical}] \gets d_{\textbf{m}_{n} \prec \textbf{m}}^{\text{vertical}}$\\
                    Queue.insert($I_{\textbf{m}_{n}}$)\;
                }
            }
        }
    }
\end{algorithm}

\begin{algorithm}[ht!]
    \caption{Determine whether $I_{\textbf{m}_b}$ is supported by $I_{\textbf{m}_a}$ based on the content of the image $I$.}
    \label{Alg:supported_relationship_analysis}
    \KwIn{Mask images $I_{\textbf{m}_a}$ and $I_{\textbf{m}_b}$, along with their corresponding oriented bounding boxes (OBBs) $\text{obb}_{\text{m}_a}$ and $\text{obb}_{\text{m}_b}$.;}
    \KwResult{Support relationship between $I_{\textbf{m}_a}$ and $I_{\textbf{m}_b}$: $S_{\textbf{m}_a \prec \textbf{m}_b}$; the relative vertical distance between $\text{obb}_{\textbf{m}_a}$ and $\text{obb}_{\textbf{m}_b}$: $d_{\textbf{m}_a \prec \textbf{m}_b}^{\text{vertical}}$;}

    \tcp{Analyze the supporting relationship.}
    \If{$ | z(\text{obb}_{\textbf{m}_{a}})_{\text{max}} - z(\text{obb}_{\textbf{m}_{b}})_{\text{min}}| < \epsilon $}{
        $S_{\textbf{m}_a \prec \textbf{m}_b} \gets \text{true}$;\\
        $d_{\textbf{m}_a \prec \textbf{m}_b}^{\text{vertical}} \gets 0$;\\
        \Return $S_{\textbf{m}_a \prec \textbf{m}_b}, d_{\textbf{m}_a \prec \textbf{m}_b}^{\text{vertical}}$
    }

    \tcp{Check if the internal relationship is satisfied.}
    \If{$\text{obb}_{\textbf{m}_{b}} \subseteq \text{obb}_{\textbf{m}_{a}}$}{
        $S_{\textbf{m}_a \prec \textbf{m}_b} \gets \text{true}$;\\
        $d_{\textbf{m}_a \prec \textbf{m}_b}^{\text{vertical}} \gets \frac{(z(\text{obb}_{\textbf{m}_{b}})_{\text{max}} + z(\text{obb}_{\textbf{m}_{b}})_{\text{min}})/2 - z(\text{obb}_{\textbf{m}_{a}})_{\text{min}}}{(\text{obb}_{\textbf{m}_{a}})_{h}}$;\\
        \tcp{($\text{obb}_{\textbf{m}})_{h}$ is the vertical height of the $\text{obb}_{\textbf{m}}$}
        \Return $S_{\textbf{m}_a \prec \textbf{m}_b}, d_{\textbf{m}_a \prec \textbf{m}_b}^{\text{vertical}}$
    }

    \tcp{Handle cases of excessive occlusion, analyzed based on GPT-4o prompts.}
    \If{$\text{obb}_{\textbf{m}_{b}}$ is supported by $\text{obb}_{\textbf{m}_{a}}$ as determined by GPT-4o}{
        $S_{\textbf{m}_a \prec \textbf{m}_b} \gets \text{true}$;\\
        $d_{\textbf{m}_a \prec \textbf{m}_b}^{\text{vertical}} \gets 0$;\\
        \Return $S_{\textbf{m}_a \prec \textbf{m}_b}, d_{\textbf{m}_a \prec \textbf{m}_b}^{\text{vertical}}$
    }
    
    $S_{\textbf{m}_a \prec \textbf{m}_b} \gets \text{false}$;\\
    $d_{\textbf{m}_a \prec \textbf{m}_b}^{\text{vertical}} \gets 0$;\\

    \Return $S_{\textbf{m}_a \prec \textbf{m}_b}, d_{\textbf{m}_a \prec \textbf{m}_b}^{\text{vertical}}$
\end{algorithm}

\textit{Step 2: Analysis of Ceiling-Supported Objects}.  
We apply GPT-4o's prompting mechanism to identify objects supported by the ceiling, creating a set of ceiling-supported objects $\{I_{\textbf{m}_{i}} | i \in \text{C}\}$. These objects typically exhibit singular logical relationships in our experiments.

\textit{Step 3: Analysis of Objects Against Structural Elements}.  
We use GPT-4o to determine which objects contact walls, yielding a set $\{\textbf{m}_{i} | i \in \text{W}\}$. We then analyze the distance from each object's oriented bounding box $\text{obb}_{\textbf{m}_i}$ to specific structural planes using $d(\text{obb}_{\textbf{m}_{i}}, (n^{w}, t^{w}))$, resulting in sets of objects against specific walls $\{\textbf{m}_{i} | i \in \text{Wall}_{w}, w \in W_{\text{total}}\}$, where $W_{\text{total}}$ denotes all walls.

For regions without clear logical relationships (set $\{S_{\text{q}}\}$), we exclude these areas to enhance scene layout controllability, updating the foreground region to $S_{\text{fg}} = S_{\text{fg}} \setminus S_{\text{q}}$.

\paragraph{Refinement of OBBs}
Occlusions between objects result in incomplete depth maps from DepthAnything-V2. As shown in Fig.~\ref{fig:scene_graph}.(a), the cabinet obscured by the table has an inaccurate depth-derived OBB. Using the floor's simple structure as reference, we correct foreground object OBBs based on scene graph relationships. For floor-supported objects like $I_{\textbf{m}_{a}}$, we ensure their OBBs maintain perpendicular relationships with the floor plane $(n_{f}, t^{f})$ and extend them to make proper contact—as illustrated by the cabinet's corrected OBB in Fig.~\ref{fig:scene_graph}.(b). This approach significantly improves spatial accuracy in the final layout.

\subsubsection{3D Asset Retrieval} \label{subsubsec:3d_asset_retrieval_details}
For each mask region $I_{\textbf{m}_{i}}$ in the scene image, our goal is to identify the most suitable 3D asset $\text{obj}_{\textbf{m}_{i}}$ from the predefined asset library $A$. Specifically, for a given mask $I_{\textbf{m}_{i}}$ and its associated category $\text{cate}_{i}^{\text{g}}$, we first utilize the inverse mapping $\mathcal{M}^{-1}$ of the category merge map $\mathcal{M}$ related to the 3D assets (see Sec.~\ref{sec:method:scene_image_parsing}) to obtain the relevant set of categories in $A$, denoted as $\{\text{asset}_{\textbf{m}_{i}}^{A}\} = \mathcal{M}^{-1}(\text{cate}_{i}^{\text{g}})$. Subsequently, we match the most similar 3D asset within the subset $\{\text{asset}_{\textbf{m}_{i}}^{A}\}$ of the 3D asset library. In particular, we define the similarity between the mask $I_{\textbf{m}_{i}}$ and the rendered images of the assets based on the semantic similarity of visual features, which in turn informs the similarity between the mask and the assets. Furthermore, inspired by HOLODECK, we introduce an absolute size difference to adjust the matching similarity, aiming to address challenging scenarios, such as those involving significant occlusion (e.g., a bedside table obstructed by a bed).

\begin{equation}
\label{eq:matching assets}
\text{match}(\text{asset}_{\textbf{m}_{i}}^{A}, I_{\textbf{m}_{i}}) = \frac{\sum_{v\in V}\text{sim}_{\text{cls}}(I_{\textbf{m}_{i}}, \mathcal{R}(\text{asset}_{\textbf{m}_{i}}^{A}, v))}{\text{Num}(V)} - \alpha \Delta S,
\end{equation}

$$
\label{eq:image_similarity}
\text{sim}_{\text{cls}}(\mathcal{R}(\text{asset}_{\textbf{m}_{i}}^{A}, v), I_{\textbf{m}_{i}}) = \cos \!\left\langle\! F_D(\mathcal{R}(\text{asset}_{\textbf{m}_{i}}^{A}, v)), F_D(I_{\textbf{m}_{i}})\! \right\rangle,
$$

$$\Delta(S) = |\frac{l_{\text{asset}_{\textbf{m}_{i}}^{A}}}{h_{\text{asset}_{\textbf{m}_{i}}^{A}}} - \frac{l_{\textbf{m}_{i}}}{{h_{\textbf{m}_{i}}}}| + |\frac{w_{\text{asset}_{\textbf{m}_{i}}^{A}}}{h_{\text{asset}_{\textbf{m}_{i}}^{A}}} - \frac{w_{\textbf{m}_{i}}}{{h_{\textbf{m}_{i}}}}|.$$

Here, $\text{sim}_{\text{cls}}(\cdot)$ denotes the cosine similarity computed between two high-dimensional feature vectors. The feature map $F_D(\cdot)$ represents the last hidden layer features extracted by the original DINOv2 model. $I_{\textbf{m}_{i}}$ is the image corresponding to the mask, and $\mathcal{R}(\text{asset}_{\textbf{m}_{i}}^{A}, v)$ is the rendered image of the asset $\text{asset}_{\textbf{m}_{i}}^{A}$, obtained from a specific viewpoint $v$ using the camera intrinsic parameters $K$.

The viewpoint $v$ corresponds to an extrinsic parameter matrix $[R^{v}| t^{v}]$. We uniformly sampled 20 viewpoints along the central axis of the wrapped regular dodecahedron of the asset, denoted as $V$. The term $\Delta S$ represents the average absolute difference between the estimated dimensions and the actual dimensions of the model. The parameters $l_{\text{asset}_{\textbf{m}_{i}}^{A}}$, $w_{\text{asset}_{\textbf{m}_{i}}^{A}}$, and $h_{\text{asset}_{\textbf{m}_{i}}^{A}}$ correspond to the length, width, and height of the asset $\text{asset}_{\textbf{m}_{i}}^{A}$, respectively. In contrast, $l_{\textbf{m}_{i}}$, $w_{\textbf{m}_{i}}$, and $h_{\textbf{m}_{i}}$ represent the length, width, and height outputs generated by GPT-4o for $I_{\textbf{m}_{i}}$ through a prompt. In our experiments, we set the parameter $\alpha$ to 0.1.

\subsubsection{Scale Transformation} \label{subsubsec:scale_transformation_details}
In the task of 3D scene layout, designers adjust the scale and proportions of asset models according to the specific requirements of the current scene. On one hand, it is crucial to ensure that the dimensions of the asset models align with the overall layout design after placement; on the other hand, the "unique design" characteristics of the original assets must be preserved. For instance, a TV cabinet can be scaled in all three dimensions of its oriented bounding box (OBB), while a floor lamp is typically scaled only in the vertical direction, with the horizontal dimensions maintaining proportional scaling. This differentiated scaling approach for various assets not only takes the global layout into significant consideration but also preserves the inherent design characteristics of each asset. We primarily optimize the scale transformation based on the OBB of the objects, as illustrated in Eq.~\ref{eq:scale_transformation}. 
\begin{equation}
    \small
    \label{eq:scale_transformation}
    s_{\text{best}} = \arg\max_{s}\text{V}(\text{obb}_{\textbf{m}_{i}} \cap \text{obb}_{\text{obj}_{\textbf{m}_{i}}}\!(s)) - \text{V}(\text{obb}_{\textbf{m}_{i}} \cup \text{obb}_{\text{obj}_{\textbf{m}_{i}}}\!(s))
\end{equation}

where $\text{V}(\cdot)$ denotes the operator that computes the volume of a geometric body, $\text{obb}_{\text{m}_{i}}$ is the oriented bounding box corresponding to the mask image $I_{\textbf{m}_{i}}$ in the scene image, and $\text{obb}_{\text{obj}_{\textbf{m}_{i}}}(s)$ corresponds to the oriented bounding box of the retrieved asset $\text{obj}_{\textbf{m}_{i}}$ with the scale variable $s$.

The optimization strategy for the scale transformation $s$ of $\text{obb}_{\text{obj}_{\textbf{m}_{i}}}(s)$ is primarily informed by the habitual layout practices of professional artists, and it analyzes the following three scenarios:

\begin{enumerate}
    \item The scale transformation $s$ has two degrees of freedom. Objects maintain their original length-to-width ratio while height can be adjusted independently. This mode is ideal for items where height modification doesn't affect aesthetic quality, such as decorative objects and lighting fixtures.
    
    \item The scale transformation $s$ has two degrees of freedom. Objects are scaled along their two longest oriented bounding box axes, with the third dimension scaled proportionally based on the average of the other dimensions. This approach works well for slender objects like picture frames, wooden boards, and curtains.
    
    \item The scale transformation $s$ has three degrees of freedom. Objects can be freely scaled in all dimensions—length, width, and height. This mode is appropriate for furniture pieces like tables, cabinets, and beds that can be proportionally adjusted in any direction.
\end{enumerate}

Based on these scenarios, we classified the assets and derived the scale transformation for the foreground objects based on Eq.\ref{eq:scale_transformation}.

\subsubsection{Global Translation Optimization} \label{subsubsec:Two_step_translation_optimization}
To analyze the solution of Eq.~\ref{eq:refined_pose_via_logstic}, we divide the solving process into two distinct steps:

\textbf{Step 1: Hard Constraint Processing.} To ensure that the solution adheres to certain hard constraints, we perform preliminary processing of the support and wall constraints. Specifically, for the support constraints, we utilize the support tree $\mathcal{T}$ established in Sec.~\ref{sec:method:scene_image_parsing}. Starting from the root node, we sequentially update the $z$ values of the child node objects according to the support constraints, ensuring that these $z$ values are not optimized in subsequent stages. For the object $\text{obj}_{\textbf{m}_i}$ that is adjacent to the wall object $\text{obj}_{w}$, we only need to move $\text{obj}_{\textbf{m}_i}$ along the direction of the normal vector $n^{w}$ to a position that is in close contact with the wall object $\text{obj}_{w}$. Furthermore, in the subsequent optimization, the position of $\text{obj}_{\textbf{m}_i}$ will not be adjusted along the direction of $n^{w}$.

\textbf{Step 2: Nonlinear Optimization.} The variables corresponding to the support and wall constraints in Eq.~\ref{eq:refined_pose_via_logstic} have changed as a result of the updates from the first step. For instance, the $z$ values of objects that satisfy the support relationships are no longer subject to optimization. If $\text{obj}_{\textbf{m}_i}$ is adjacent to the wall object $\text{obj}_{w}$, then the change of $\text{obj}_{\textbf{m}_i}$ in the direction of $n^{w}$ is zero. However, the remaining problem still constitutes a highly nonlinear optimization challenge, which we address using a simulated annealing algorithm. Additionally, to accelerate the convergence of the solution, we employ a voxel representation of the objects as a proxy for the polygon mesh, significantly reducing the computational complexity associated with intersection calculations.

\subsubsection{Scene Layout Similarity Function}
\label{subsubsection:Scene_Layout_Similarity_Metric}
\begin{figure*}[!htbp]
    \centering
    \includegraphics[trim=0 10 0 5, clip, width=1\textwidth]{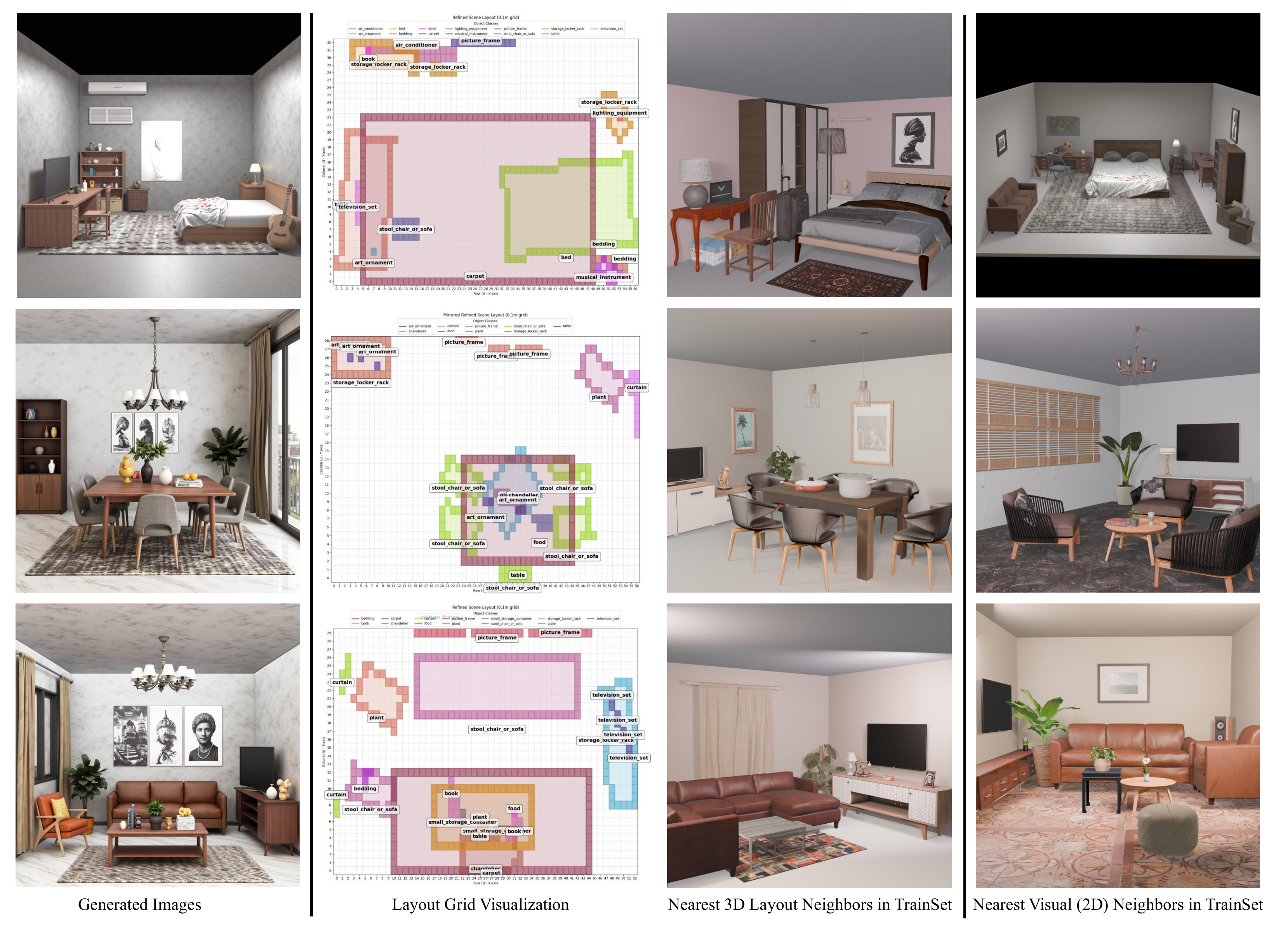}
    \caption{Example scene visualizations with 3D layout and 2D visual similarity comparisons. The first column shows generated scene images. The second column displays their corresponding grid-based layout visualizations, with each color representing a different furniture category. The third column presents the nearest neighbors based on 3D layout similarity, and the fourth column shows the nearest visual (2D) neighbors from the training set.
    }
    \label{fig:layout_grid_vis}
\end{figure*}

To quantitatively evaluate layout similarities between two scenes, we implemented a scene-to-scene similarity function following prior works. For each generated scene, we first applied our scene image analysis pipeline (Section \ref{sec:method:scene_image_parsing}) to obtain semantic segmentation results, point cloud oriented bounding boxes (OBB), and the center and plane equations of walls and floors. With this information, we project all objects onto the floor plane, aligning the scene orientation with the wall direction to standardize the coordinate system (aligning with either the x-axis or y-axis). We then construct a grid with each cell measuring 0.1m $\times$ 0.1m. For each grid point, we record the class of the furniture item present, or 'none' if empty. To enable direct comparison, we pad both scenes to the same dimensions before calculation. The similarity between a ground-truth room and a generated sample is calculated as the fraction of grid points with matching classes. To ensure comprehensive comparison, we allow for rotations of 90°, 180°, and 270° degrees, as well as mirroring transformations, to identify the maximum layout similarity between the compared scenes. Fig.~\ref{fig:layout_grid_vis} shows several example scenes.

\subsubsection{More Details in Physical Constraints}
\label{subsubsec:More_Details_in_Physical_Simulation}
Table~\ref{tab:physical_simulation} summarizes the simulation parameters used in Blender for physics simulations.

\begin{table*}[htbp!]
    \centering
    \caption{Summary of simulation parameters used in Blender for physics simulation.}
    \label{tab:physical_simulation}
    \resizebox{0.75\textwidth}{!}{ 
    \begin{tabular}{llll}
        \toprule
        \multicolumn{4}{c}{\textbf{Simulator Parameters}} \\
        \midrule
        scene.frame\_start & 1 & scene.rigidbody\_world.solver\_iterations & 3 \\ 
        scene.frame\_end & 200 & scene.rigidbody\_world.substeps\_per\_frame & 3 \\ 
        scene.gravity & (0,0,-9.81) & & \\
        \midrule
        \multicolumn{4}{c}{\textbf{Rigid Body Simulation Parameters}} \\
        \midrule
        obj.rigid\_body.mass & 10 & obj.rigid\_body.collision\_shape & MESH \\ 
        obj.rigid\_body.friction & 10 & obj.modifiers & Decimate-DECIMATE \\ 
        obj.rigid\_body.restitution & 0 & modifier.decimate\_type & DISSOLVE \\ 
        obj.rigid\_body.linear\_damping & 1 & modifier.angle\_limit & 15 degrees \\ 
        & & obj.rigid\_body.collision\_margin & 0.001 \\ 
        & & obj.rigid\_body.use\_deform & TRUE \\ 
        \bottomrule
    \end{tabular}
    }
\end{table*}

\subsubsection{Additional Visualization of Intermediate Results}
\label{subsubsec:supp_vis}
\begin{figure*}[!htbp]
    \centering
    \includegraphics[trim=0 0 0 0, clip, width=0.95\textwidth]{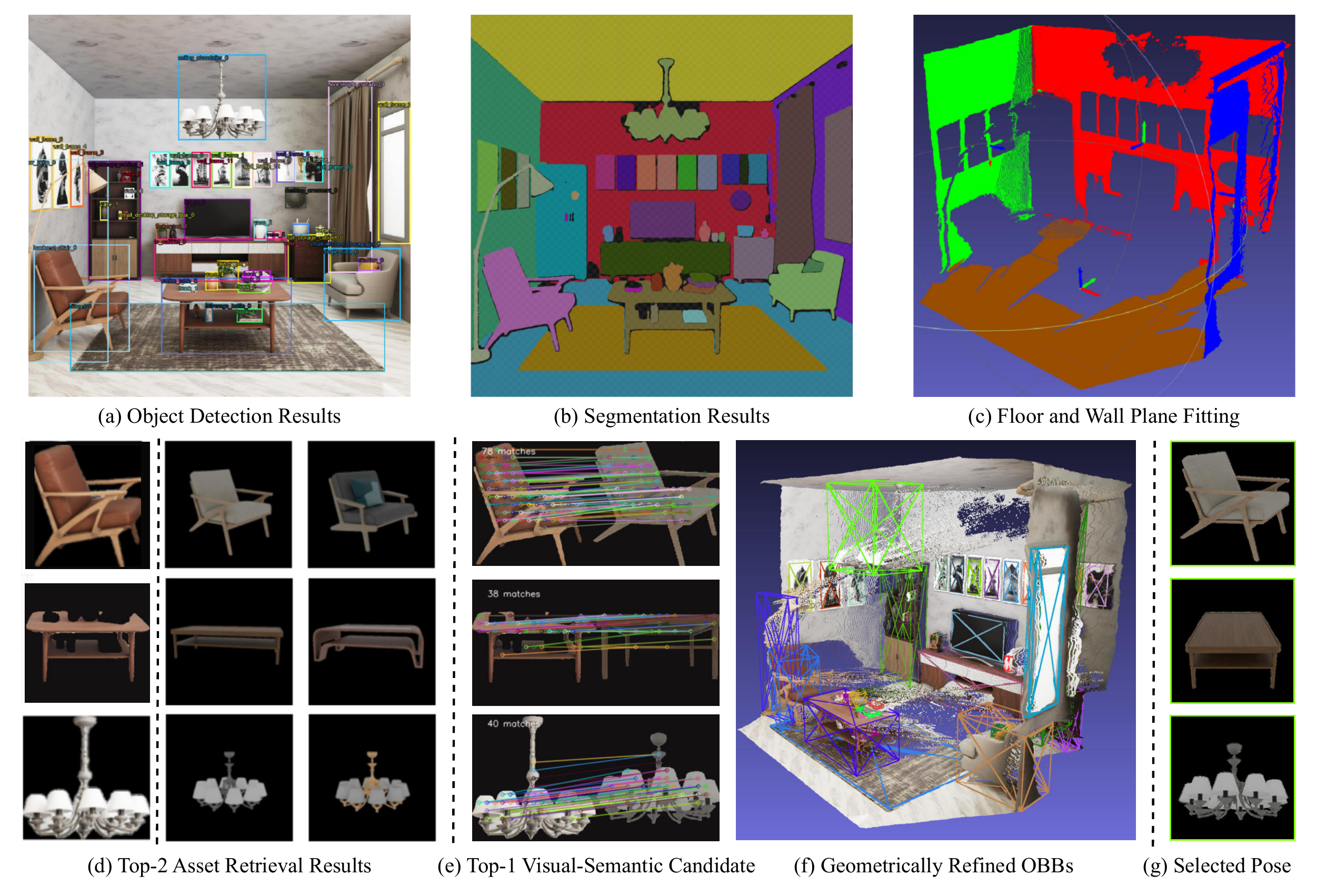}
    \caption{Additional visualization of key intermediate steps. The process begins with a comprehensive scene analysis where (a) objects are detected via grounding-dino-1.5 and SAM, guided by categories parsed by GPT-4o, and (b) segmentation masks are generated. Concurrently, (c) RANSAC is employed to fit orthogonal floor and wall planes (ceiling as floor's opposite normal), establishing a robust geometric frame for the scene. For each segmented object, we (d) retrieve the top-2 candidate assets from our library based on semantic category, visual similarity, and size compatibility. Our rotation estimation module then combines (e) a strong initial candidate from visual-semantic feature matching with (f) constraints from Oriented Bounding Boxes (OBBs), which are geometrically corrected using scene graph logic. This fusion results in (g) the final pose, a high-quality input for the subsequent scene layout refinement stage.
    }
    \label{fig:supp_inter_res_vis}
\end{figure*}

To offer a comprehensive visual overview of our method, we present visualizations from different stages of our pipeline. Fig.~\ref{fig:pipeline} illustrates the overall workflow, showcasing the initial guide image and the layout results before and after the final layout refinement. Fig.~\ref{fig:scene_graph} details the constructed scene graph. Complementing these, Fig.~\ref{fig:supp_inter_res_vis} provides a granular breakdown of the intermediate algorithmic details, demonstrating the process from scene parsing and asset retrieval to the final rotation estimation for individual objects.

\subsection{Prompts} \label{subsec:prompts}
\subsubsection{Complete scene generation prompts}
\label{subsubsec:Complete_scene_generation_prompts}
The following are the complete text prompts used to generate the scenes shown in Fig.~\ref{fig:teaser}:

\medskip
\styledfileinput[python]{sec/tabs/supplementary/complete_prompts.txt}{}

\subsubsection{Prompt for Object Extraction}
\label{subsubsec:Prompt_for_Object_Extraction}
The following prompt is designed to extract all objects within a scene using a Chain-of-Thought (CoT) approach. The output is formatted as a JSON object list.
\medskip
\styledfileinput[python]{sec/tabs/supplementary/prompt2.txt}{}

\subsubsection{Prompt for Scene Layout Analysis}
\label{subsubsec:Prompt_for_Scene_Layout_Analysis}
The following prompt is specifically crafted for analyzing the structural dependency relationships among objects in a structured and hierarchical manner. The analysis follows stringent guidelines and produces results in a JSON format.
\medskip
\styledfileinput[python]{sec/tabs/supplementary/prompt3.txt}{}

\begin{figure*}
    \centering
    \includegraphics[trim=0 240 0 0, clip, width=0.9\textwidth]{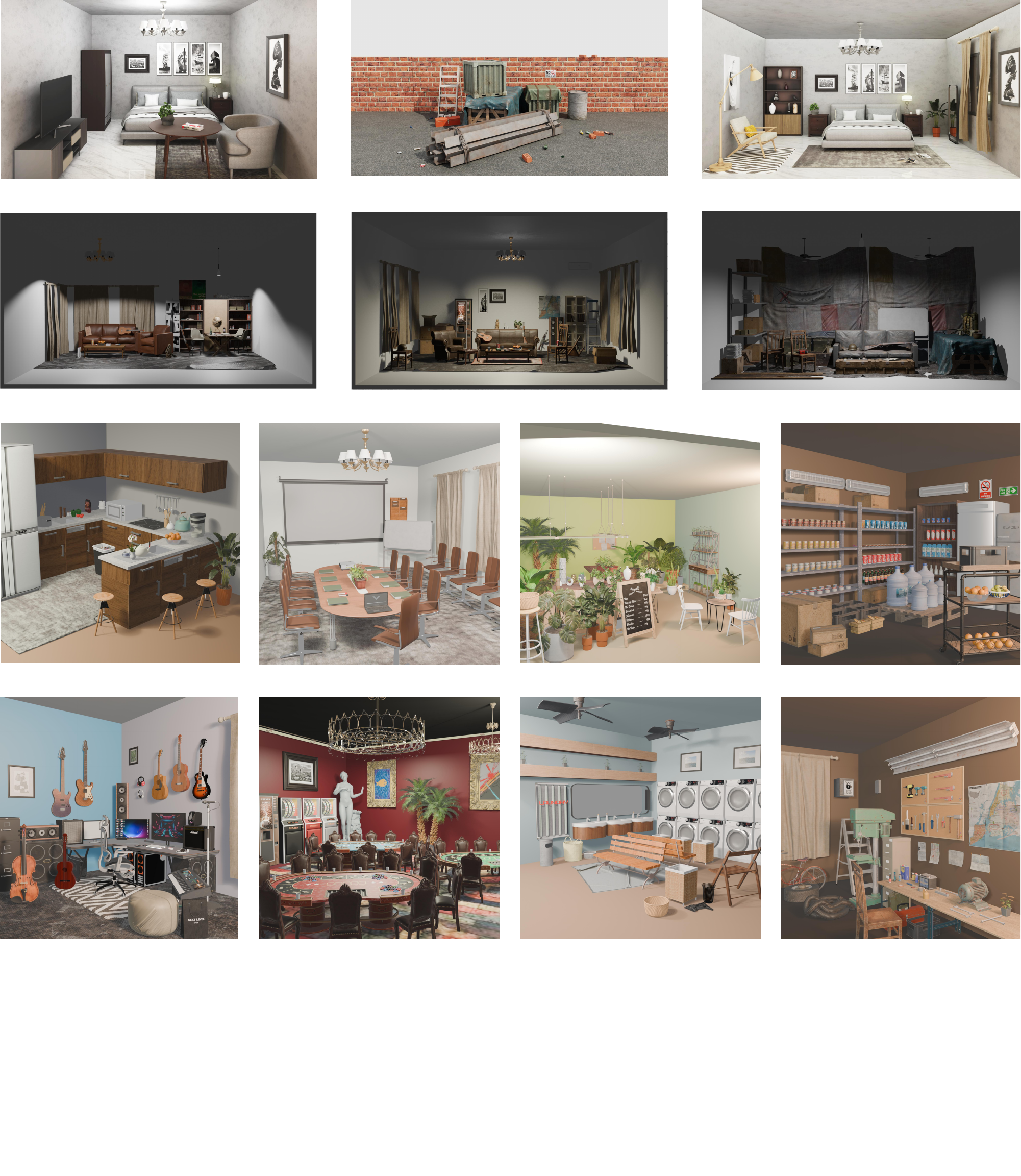}
    \caption{More 3D scenes from our high-quality, handcrafted dataset. These scenes showcase a diverse range of room functions and include both indoor and outdoor assets, illustrating the variety and detail of our manual scene construction.}
    \label{fig:dataset_catpion}
\end{figure*}

\subsubsection{Prompt for GPT4 evalutation}
\label{subsubsec:Prompt_for_GPT4_evalutation}
Below are the prompts used in expert evaluation and layout similarity assessment experiments. We utilized the GPT-4 model and set the temperature to 0.
\medskip 
\styledfileinput[python]{sec/tabs/supplementary/prompt1_1.txt}{}
\medskip
\styledfileinput[python]{sec/tabs/supplementary/prompt1_2.txt}{}

\subsection{Dataset Details} \label{subsec:More_Details_for_Dataset}
\begin{figure*}[t!]
    \centering
    \includegraphics[trim=215 5 215 5, clip, width=\textwidth]{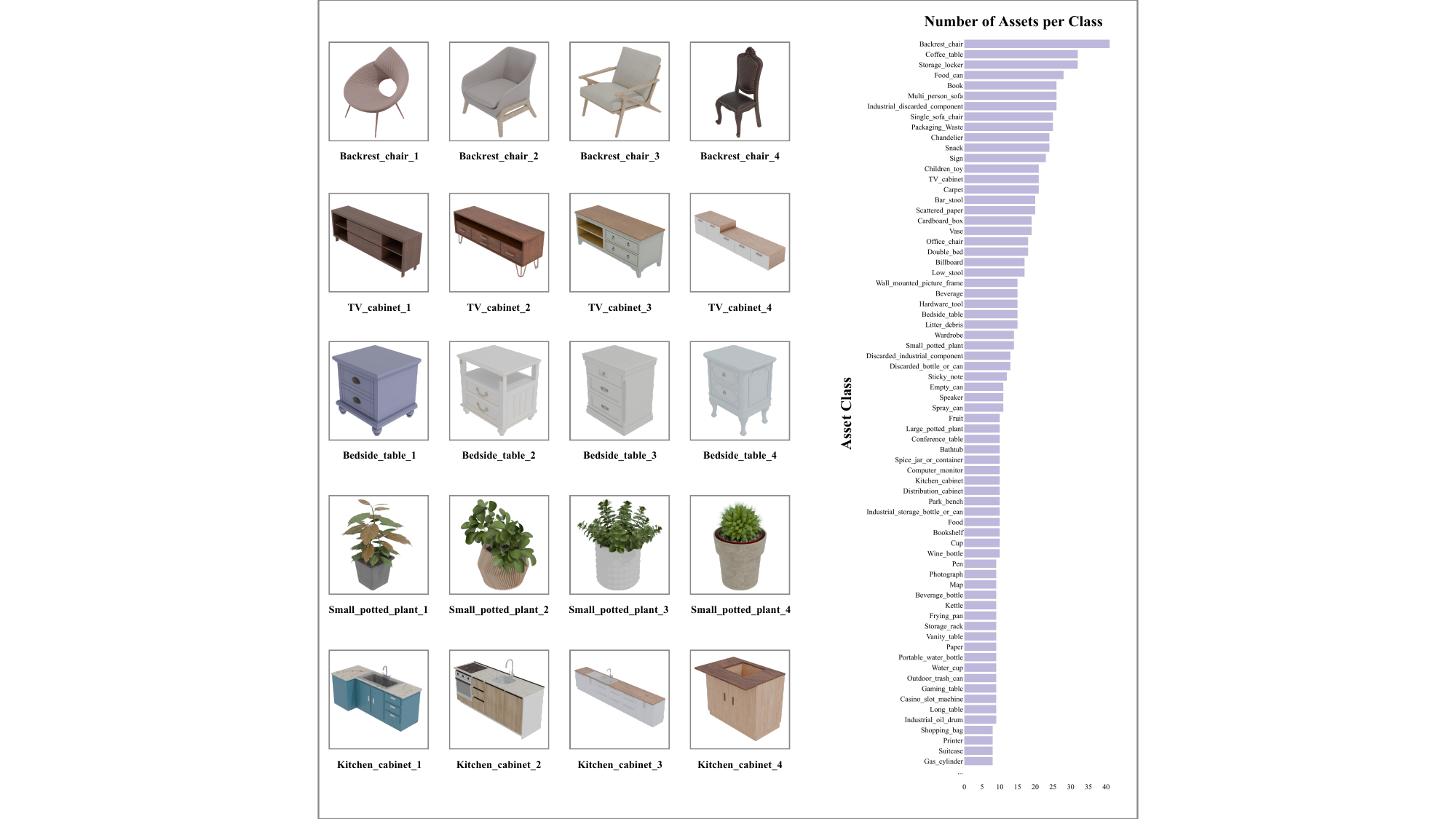}
    \caption{Dataset asset overview. Left: examples of asset classes such as backrest chairs and TV cabinets. Right: bar chart showing the number of assets per class, highlighting the most common categories.}
    \label{fig:asset_overview}
\end{figure*}

\begin{figure*}[t!]
    \centering
    \includegraphics[trim=0 0 0 0, clip, width=0.90\textwidth]{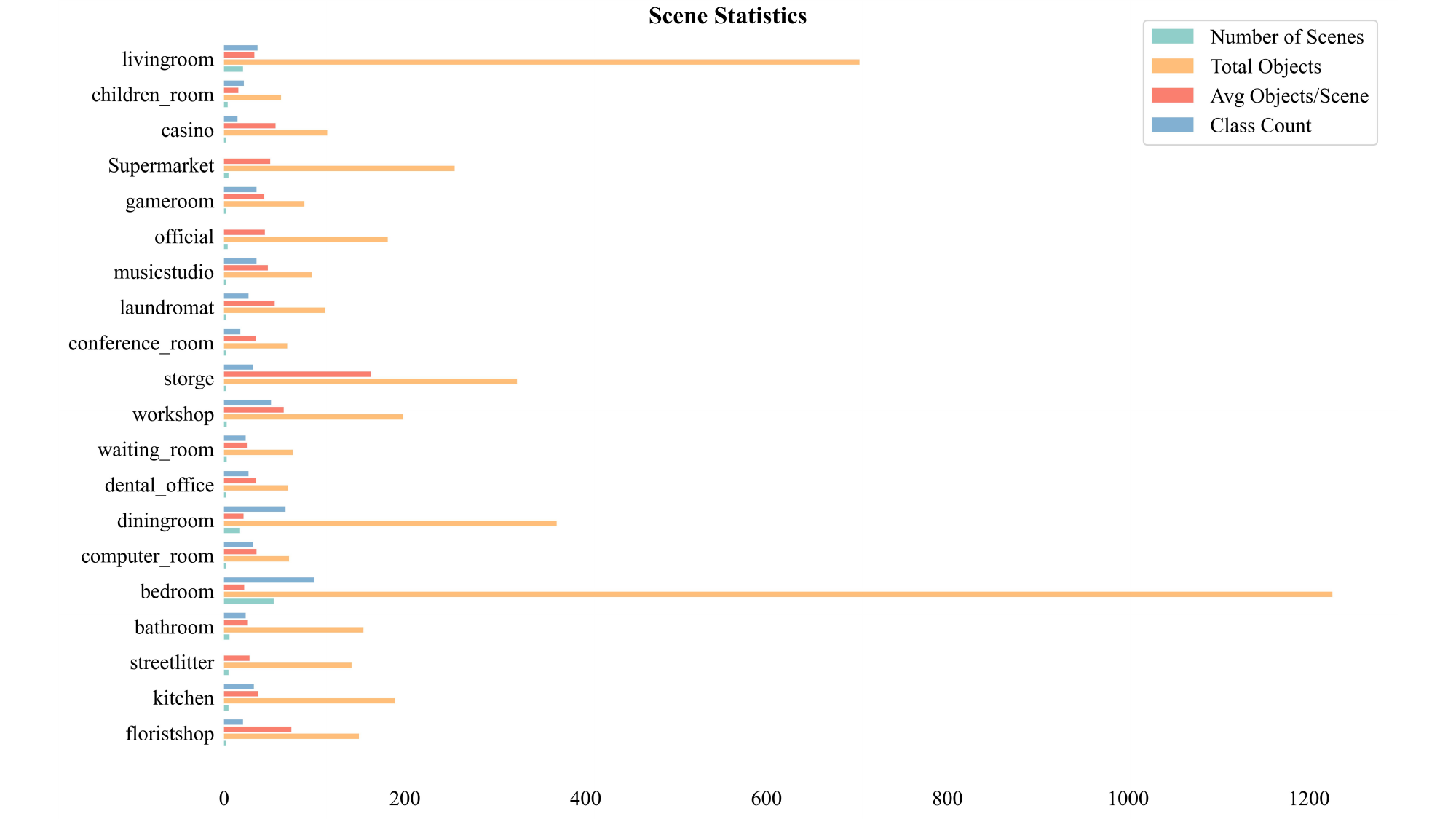}
    \caption{Statistics of our high-quality, manually arranged dataset, encompassing 21 different scene types. The chart illustrates the number of scenes, total objects, average objects per scene, and class count for each scene type, highlighting the dataset's diversity and complexity.}
    \label{fig:Scene_statistics}
\end{figure*}

Our dataset addresses several key limitations of existing 3D scene layout resources, significantly enhancing both quality and diversity. As illustrated in Fig.~\ref{fig:dataset_catpion}, we present a variety of high-quality, handcrafted 3D scenes that showcase diverse room functions. Fig.~\ref{fig:asset_overview} showcases examples of our diverse assets, with the left side displaying representative high-quality models and the right side presenting a bar chart that illustrates the distribution across various categories. For common items, we offer multiple variants to capture different styles. Additionally, as shown in Fig.~\ref{fig:Scene_statistics}, we provide a statistical analysis of the number and distribution of object types in a sample of scenes.

\subsection{Analysis of Failure Cases}
\label{subsec:failure_analysis}
\begin{figure*}[t!]
    \centering
    \includegraphics[trim=0 0 0 0, clip, width=0.95\textwidth]{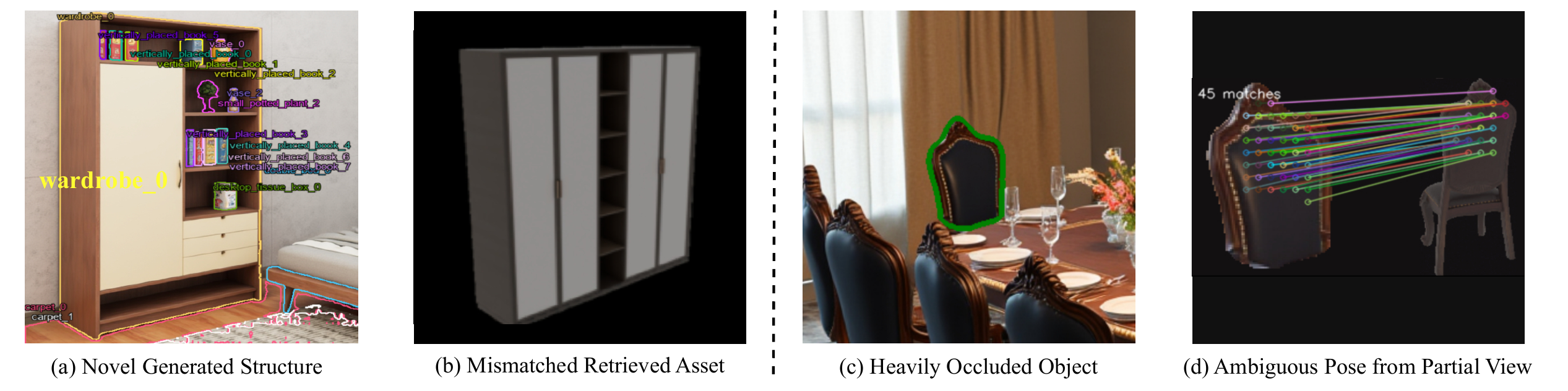}
    \caption{Analysis of Failure Cases. This figure illustrates two primary limitations of our method. Top Row: Discrepancy between Generated Content and Asset Library. (a) The image generator creates an object with a novel topology—a hybrid of a wardrobe and a bookshelf. (b) Our system retrieves the closest semantic match from the asset library, a standard wardrobe, which lacks the open shelves depicted. This semantic-structural mismatch prevents the correct placement of child objects (e.g., books), leading to layout inconsistencies. Bottom Row: Pose Estimation Ambiguity from Severe Occlusion. (c) An object, correctly identified as a chair, is heavily occluded, revealing only its backrest. (d) While feature matching can be performed on this partial view, the limited information introduces ambiguity, as multiple poses could yield a similar appearance, potentially leading to inaccurate rotation estimation.
    }
    \label{fig:failure_cases}
\end{figure*}

The failure cases illustrated in Fig.~\ref{fig:failure_cases} highlight two core challenges. A semantic-structural mismatch can occur when the image generator produces objects with novel topologies not present in our finite asset library (top row). This leads to incorrect asset retrieval, which in turn invalidates downstream geometric and relational constraints derived from the scene graph. Furthermore, pose ambiguity from severe occlusion remains a key limitation (bottom row). As an inherently ill-posed problem, the partial view from an occluded object provides ambiguous visual features for our matching module, leading to an unreliable initial pose estimate that subsequent optimization stages may fail to correct.

\subsection{More Qualitative Results}
\label{subsec:More_Qualitative_Results}
\begin{figure*}[t!]
\centering
\includegraphics[trim=0 10 0 0, clip, width=1\textwidth]{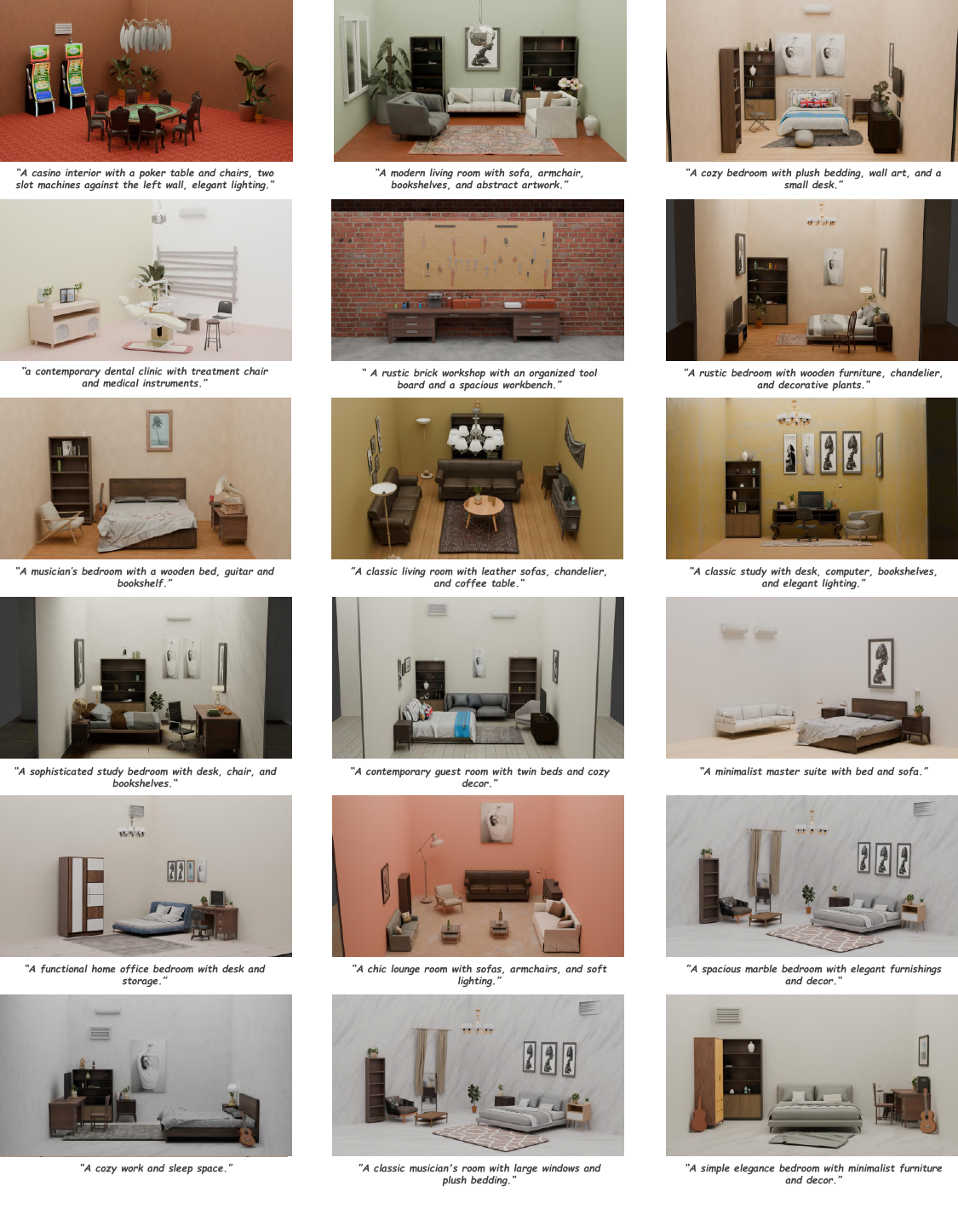}
\caption{Additional 3D generated scene layouts by our system.}
\label{fig:additional_scene}
\end{figure*}

To further demonstrate the ability of our algorithm to generate diverse 3D scene layouts, we present additional 3D scenes produced by our method in Fig.~\ref{fig:additional_scene}.

\clearpage

\end{document}